%% file: neurips_2026.tex
\documentclass{article}

\PassOptionsToPackage{numbers, sort}{natbib}

\usepackage[dblblindworkshop, final]{neurips_2026}
\usepackage{graphicx}
\usepackage{rotating}

\workshoptitle{TAE (Trust-AI-Eval): Can We Trust AI Evaluation?}

\usepackage[utf8]{inputenc}
\usepackage[T1]{fontenc}
\usepackage{hyperref}
\usepackage{url}
\usepackage{booktabs}
\usepackage{amsmath}
\usepackage{float}
\usepackage{algorithm}
\usepackage{algpseudocode}
\usepackage{amsfonts}
\usepackage{nicefrac}
\usepackage{microtype}
\usepackage{xcolor}
\usepackage{listings}
\usepackage[dvipsnames]{xcolor}
\usepackage{wrapfig}
\usepackage{array}

\title{PADM\'E: Preference Alignment Data Synthesis for Meta-Evaluation of LM Agent Evaluators}

\author{%
  Cheng Chang \\
  Uniphore \\
  \texttt{cheng.chang@uniphore.com} \\
  \And
  Yining Mao \\
  Uniphore \\
  \texttt{yining.mao@uniphore.com} \\
  \AND
  Peng Qi \\
  Uniphore \\
  \texttt{peng.qi@uniphore.com} \\
}

\begin{document}

\maketitle

\begin{abstract}
Language models are frequently employed to evaluate other language models.
An LM evaluator scoring agentic behaviors across multiple criteria is valuable, provided that its decisions align with human judgment.
We call the problem of evaluating this alignment \textbf{Meta-Evaluation}.
Tackling it directly is difficult: collecting human data is expensive, absolute scoring is hard to align, and using an LM meta-evaluator recurses the question of trustworthiness.
We adopt a reformulation of meta-evaluation as a preference judgment problem: rather than comparing human and LM evaluator scores of a trajectory, we ask whether their \textit{implied preferences} align.
Building on this, we introduce PADM\'E, a data synthesis method that generates reliable criterion-based meta-evaluation data for agentic settings.
PADM\'E uses only small language models, requires no human involvement during evaluations, and operates under a low computational budget.
We build a prototype of PADM\'E and synthesize a dataset of 1{,}000 samples across four agentic domains and three evaluation criteria.
Human validation on a 150-sample subset demonstrates that PADM\'E improves agreement with human judgment from 73\% to 85\% over a naive baseline.
Meta-evaluating 25 common models with our dataset demonstrates the correlations between evaluation performance and scoring granularity, leniency, and model size, among other factors.

\end{abstract}

\section{Introduction}
\label{sec:introduction}

Language models (LMs) are routinely used to evaluate LM agent systems\footnote{Here, we broadly define an LM agent system as a computer program that uses language models to hold multi-turn conversations with users, call tools (other computer programs) to aid its work, and change the state of a virtual environment on behalf of its users.
Critically, it produces a trajectory rather than a single response during one user session.} in research and commercial use cases \citep{mtbench, zhuge2025agentasajudge, judgesurvey}, and the evaluations are frequently applied over a set of distinct performance metrics \citep{ye2024flask, kim-etal-2025-biggen, bai-etal-2024-mt}.
For example, the developers of a customer-service agent on a shopping website would reasonably want to know how the agent scores, across thousands of conversations, on evaluation criteria such as friendliness, factuality, and answer relevance.
The evaluation criteria can be stylistic, open-ended, or demand an intelligent understanding of whole conversations and their outcomes to be properly evaluated, so the environment lacks natural and deterministic signals for assessing them \citep{zhuge2025agentasajudge, DBLP:conf/uist/ShankarZHPA24}.
In these cases, LM evaluators may be the only viable option for automated evaluations.

\begin{figure}
\includegraphics[width=\textwidth,trim=1.9cm 2.94cm 2.4cm 5.9cm, clip=true]{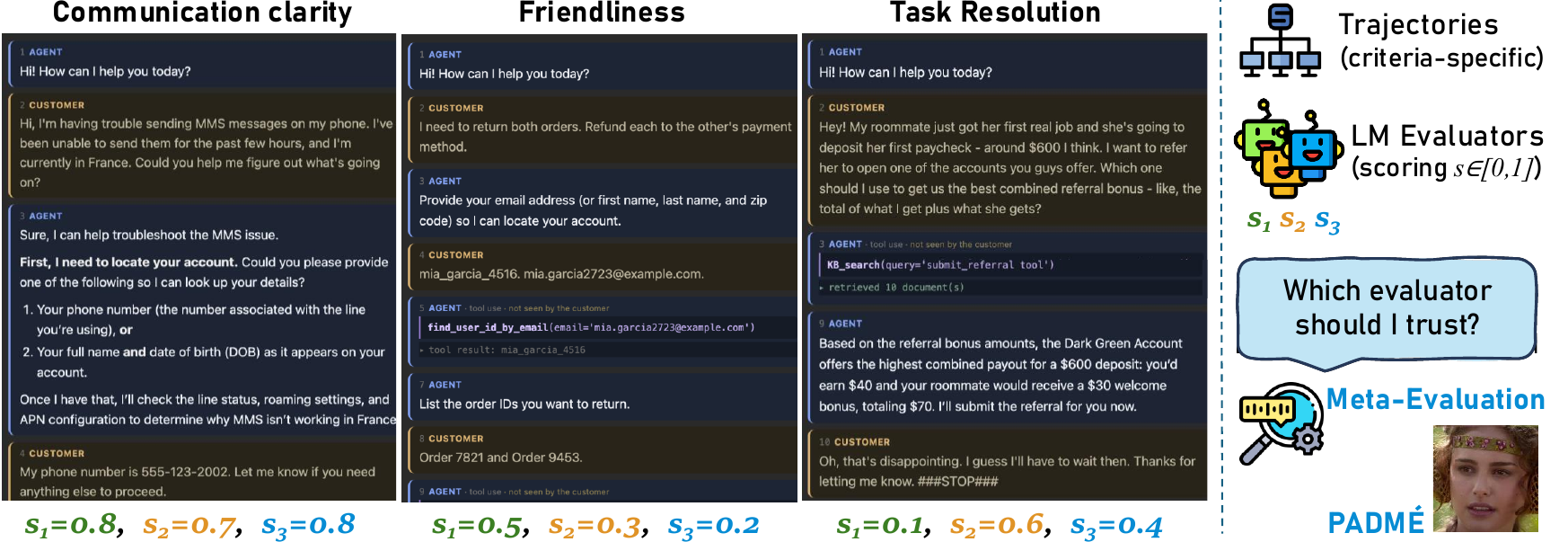}
  \caption{\textbf{Overview of criterion-specific meta-evaluation.} Given multi-step user-agent interaction trajectories evaluated under a specific criterion $c$ (e.g., communication clarity), an LM evaluator $f$ assigns a continuous score $s_f \in [0, 1]$.
  Meta-evaluation asks the question ``Which of the evaluators ($s_1, s_2, s_3$) should I trust?'', which PADM\'E attempts to solve through preference alignment.}
  \label{fig:meta-evaluation-problem}
\end{figure}

For the scores provided by the LM evaluator to be useful, they must reliably reflect the actual performance of the agent over the criteria.
For agentic systems that mostly serve human users through conversations and environment interactions, the alignment between human and LM evaluator judgments is a natural estimate of the reliability of the LM evaluator \citep{mtbench, judgesurvey, DBLP:conf/uist/ShankarZHPA24}.
We call the assessment of this alignment the \textbf{Meta-Evaluation} problem (Figure~\ref{fig:meta-evaluation-problem}).

Meta-evaluation is well established for single-turn response evaluations and reinforcement learning with human feedback \citep{rewardbench, rewardbench2, judgebench, ifrewardbench}.
It has only recently reached agentic settings in limited scenarios \citep{zhuge2025agentasajudge, ajbench, onlineagentjudge}.
A direct method would involve collecting human annotations for each evaluated agent over each criterion of interest \citep{mtbench, DBLP:conf/uist/ShankarZHPA24, rocketeval}, but human labeling is costly,
which makes it especially prohibitive for rapidly developing agent platforms with a massive number of agent designs, use cases, and datasets.
Additionally, maintaining consistency when rating items on an absolute scale is hard for human annotators \citep{bws, rankme}.
Another direct solution is to use a more capable model, a ``meta-evaluator'', to judge the evaluation of the LM evaluator.
For example, finetuned evaluators are commonly benchmarked against frontier LMs \citep{prometheus2, zhu2025judgelm, wang2024pandalm}.
However, this method requires access to stronger models, which can be prohibitive in itself.
It also recurses the meta-evaluation problem, since nothing guarantees the reliability of the LM meta-evaluator itself \citep{DBLP:conf/uist/ShankarZHPA24, judgebench}.

In this paper, we present a simple, reliable, and inexpensive method for automatic meta-evaluation of arbitrary LM evaluators.
We avoid the difficulties mentioned above with two designs.
First, instead of asking whether humans would give a trajectory the same \textit{score} as the LM evaluator does, we hand the evaluator two trajectories to score separately, and ask whether the \textit{preference} implied by the score difference agrees with the human preference (Section~\ref{sec:metaeval}).
This is an established practice in prior research (Section~\ref{sec:related}).
Second, we define an algorithm that synthesizes labeled trajectory pairs for a given agent system and an arbitrary set of criteria, using small language models (SLMs) no more capable than the ones the existing evaluator and agent system rely on (Section~\ref{sec:curation}).
No human input is needed during data synthesis or at evaluation time.

We call the algorithm \textbf{PADM\'E}, or Preference Alignment Data synthesis for Meta-Evaluation. Building on $\tau^3$-bench \citep{yao2025taubench, barres2026taubench, shi2026tauknowledge}, we develop an instance of PADM\'E and use it to synthesize 1{,}000 data pairs across four task domains and three criteria (Section~\ref{sec:produced}), validate 150 of them with six human annotators (Section~\ref{sec:human}), and evaluate 25 common models as evaluators (Section~\ref{sec:discriminates}).

We claim three contributions.
\begin{itemize}
    \item \textbf{A data synthesis algorithm}
    that builds meta-evaluation data for agent systems and any evaluation criteria, as opposed to a fixed benchmark (Section~\ref{sec:method}).
    \item \textbf{A program} that implements this algorithm for a specific use case (Section~\ref{sec:setup}), together with a synthetic dataset generated by that program (Section~\ref{sec:produced}).
    \item \textbf{Experiments} showing that the algorithm is data- and cost-efficient, and a human study showing that the preference labels this algorithm constructs agree with human judgment in this scenario (Section~\ref{sec:results}).
\end{itemize}
We release our code and the synthesized data at \url{https://github.com/chc012/padme}.

\section{Related Work}
\label{sec:related}

\paragraph{Judging Agents, and Meta-Evaluating the Judges}
LLM-as-a-judge is the paradigm for open-ended evaluation \citep{mtbench, judgesurvey}.
Increasingly, LM evaluators target multi-turn behaviors rather than single-turn responses \citep{zhuge2025agentasajudge, onlineagentjudge, ajbench}.
Meta-evaluation of these evaluators has evolved from coarse response-level preferences \citep{rewardbench, rewardbench2, judgebench} toward finer-grained units, such as skill decompositions \citep{ye2024flask}, checklists \citep{cook2024tickingboxesgeneratedchecklists, rocketeval, autochecklist}, rubrics \citep{kim-etal-2025-biggen, rubriceval}, instruction constraints \citep{ifrewardbench}, and crowdsourced criterion labels \citep{wang2024helpsteer}.
For meta-evaluation in agentic settings, Agent-as-a-Judge \citep{zhuge2025agentasajudge} and AJ-Bench \citep{ajbench} use verifier scripts and manual requirement annotations, so the label exists only in use cases where completion rules are predetermined.
They explore meta-evaluation for code generation \citep{zhuge2025agentasajudge}, search, data-system manipulation, and GUI interaction \citep{ajbench}.
Two costs remain in all of these works: labels rely heavily on human curation, and data generation relies on frontier models.

\paragraph{Constructed Quality Differences}
LLMBar \citep{llmbar} is an early instance of this idea. It releases $419$ hand-curated output pairs in which one response follows the instruction and the other deviates.
FBI
\citep{blindspots} injects hand-authored perturbations that degrade one capability and asks whether an evaluator notices.
It requires manual label verification.
RubricEval \citep{rubriceval} samples responses from a mixed model pool to elicit performance differences.
Its labels are judgments of binary questions regarding each trajectory.
REFLECT \citep{reflect} meta-evaluates judges of deep research agents.
It derives controlled perturbations of agent trajectories from a taxonomy of failure patterns.
Its pipeline requires human experts for validation.
All of these works build their data with frontier models or human annotations, which restricts extension to other agentic use cases.
We attempt to offer a more broadly applicable data synthesis recipe with only SLMs (Section~\ref{sec:setup}).

\paragraph{Preferences Versus Scores}
Eliciting pairwise preference from pointwise scores is an established practice.
Comparative elicitation recovers a latent scale from pairwise judgments in psychology and statistics \citep{thurstone1927, bradleyterry1952}.
In natural language processing, it yields more reliable human labels than rating scales \citep{bws, rankme}.
Pairwise ranking aligns LLM evaluators with human judgment better than direct scoring \citep{pairs}.
Some reward-model benchmarks rate each completion independently and compare score differences \citep{rewardbench2}.
Within LLM-as-a-judge, \citet{mtbench} convert single-answer grades into pairwise comparisons in order to measure agreement with human votes.
Consequently, we acquire ground-truth data via preferences while prompting the evaluator to score each trajectory \emph{pointwise} (Eq.~\eqref{eq:gap}), which mirrors deployment conditions.
Alternatively, many meta-evaluation studies do measure scoring alignment by correlating evaluator scores directly against human ratings \citep{geval, prometheus2, ye2024flask, kim-etal-2025-biggen}.

\paragraph{SLMs as Judges}
Deploying small language models as judges serves as a premise for our work.
SLM judges are competitive across model families and parameter scales \citep{laddha2026slmjurysmalllanguagemodels}, specialized mini-evaluators have emerged \citep{prometheus2, selenemini}, and lightweight models are well-suited for repetitive agentic sub-tasks \citep{belcak2025small}.
However, SLMs are more susceptible to assertiveness and verbosity confounds \citep{tripathi2025pairwise}.
For SLM meta-evaluation, SLMJury \citep{laddha2026slmjurysmalllanguagemodels} meta-evaluates SLM judges, but it uses existing public datasets rather than building agent- and criterion-specific data.

\section{Method}
\label{sec:method}

\begin{figure}
\includegraphics[width=\textwidth,trim=2.15cm 5.12cm 5.0cm 5.72cm, clip=true]{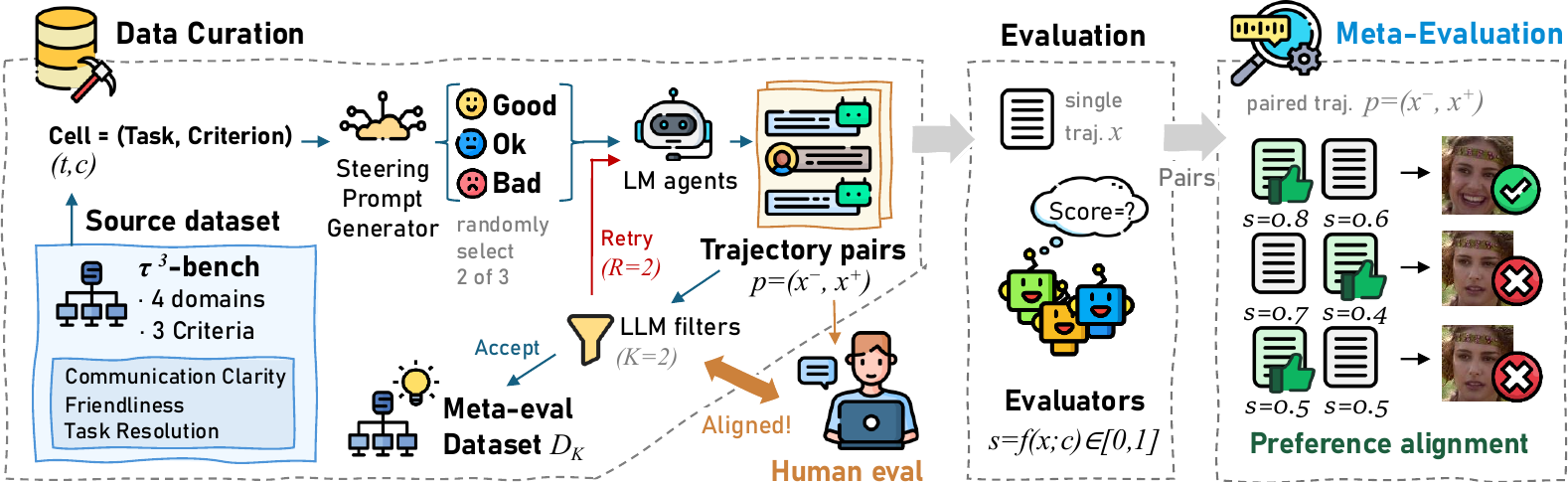}
  \caption{%
  \textbf{Overview of the PADM\'E pipeline and meta-evaluation architecture.} For each task-criterion cell $(t,c)$ from $\tau^3$-bench, steering prompts generate trajectory pairs $p = (x^-, x^+)$ across three quality levels (\textsf{good}, \textsf{ok}, \textsf{bad}). Two LM filter judges discard pairs lacking visible quality contrast to yield dataset $\mathcal{D}_K$, validated against a human annotation subset. Target evaluators rate individual trajectories ($s = f(x;c) \in [0, 1]$), and meta-evaluation assesses their preference alignment across verified pairs.}

  \label{fig:overview}
\end{figure}

As mentioned in Section~\ref{sec:introduction}, meta-evaluating an evaluator requires assessing its alignment with human judgment.
Through the pointwise-to-pairwise reframing of the evaluation objective, we try to assess whether an evaluator will rank two agent trajectories correctly by scoring them independently.
PADM\'E synthesizes trajectory pairs with known labels over arbitrary criteria through steering agent prompts.
This shifts the core challenge from annotation to data collection.
To improve label fidelity, PADM\'E applies filtering and retries after initial trajectory generation.
We define the algorithm below.
Figure~\ref{fig:overview} gives an overview of the method.

\subsection{Data Curation}
\label{sec:curation}

\paragraph{Generation}
We define the input of PADM\'E as a \textit{cell} $(t, c)$, consisting of task parameters $t$ and evaluation criterion $c$.
$t$ may include textual descriptions of the task (e.g., ``rebook a plane ticket''), tools and resources available to the agent for the task (e.g., functions \texttt{list\_purchased\_tickets} and \texttt{buy\_ticket}), information about the simulated user's intention of the task (e.g., ``if rebooking is not possible, cancel the ticket''), and so on\footnote{The selection of information available here is up to the developer of the specific agent system.
Care must be taken here to ensure that no unwanted information leakage happens inside the task parameters that would reveal the solution of the task directly to the trajectory-generating agent.}.
$c$ consists of the name and a description of the criterion. The criterion can be any performance axis of the agent system, such as ``task completion rate'', ``user satisfaction'', ``tool-use relevance'', etc.
The descriptions should ideally be detailed and align with application scenarios.
As an example, see Appendix~\ref{app:prompt_criteria} for the criteria descriptions in our experiment.
For good coverage of behavior, the source dataset should be a benchmark used to evaluate the targeted agent system, though any dataset that provides sets of $t$s and $c$s would apply.

We use each cell as a seed datum to generate a trajectory pair $p = (x^-, x^+)$ with two different steering levels, $\ell, \ell' \in \mathcal{L} = \{\textsf{bad} \prec \textsf{ok} \prec \textsf{good}\}$.
Here, a trajectory denotes a complete record of interactions between a user and an agent in a task execution session.
Trajectories are sampled using a rollout function $\mathrm{Roll}$:
\begin{equation}
x^{\ell} \sim \mathrm{Roll}\bigl(t, s_{c,t,\ell}; \theta\bigr),
\qquad
p = (x^{-}, x^{+}) := \bigl(x^{\ell}, x^{\ell'}\bigr) \quad \text{for } \ell \prec \ell'.
\label{eq:gen}
\end{equation}
The steering instruction $s_{c,t,\ell}$ is generated by a $\textsc{WriteSteer}$ function, which uses an SLM conditioned on the criterion ($c$), task ($t$), and the steering intent ($\ell$).
The instruction is then appended to the agent's system prompt (Appendix~\ref{app:prompt_wrapper}).
Each cell randomly selects two distinct steering levels, with the three resulting contrasts ($\{\textsf{bad}, \textsf{ok}\}$, $\{\textsf{ok}, \textsf{good}\}$, and $\{\textsf{bad}, \textsf{good}\}$) stratified equally across the dataset.
For example, a \textsf{bad} steering instruction on ``friendliness'' may ask the agent to be cold and concise in its response, an \textsf{ok} one to be neutral in its tone, and a \textsf{good} one to be warm and comforting in its replies.
Each instruction is tailored to the specific task and carries different information.
The generator prompt that produces these instructions is in Appendix~\ref{app:prompt_generator}.

All rollouts within a cell share a fixed environment context $\theta$, which may include the agent model, user simulator, domain configuration, and decoding seed.
While fixing $\theta$ ensures the steering instruction is the primary deliberate variable, LM stochasticity and dynamic environment simulations naturally introduce trajectory variance.

\paragraph{Filtering}
Each generated trajectory pair $p = (x^-, x^+)$ is sequentially evaluated by a cascade of $K$ SLM preference judges, $J_1, \dots, J_K$ (Appendix~\ref{app:prompt_judge}).
Given a pair $p$ in randomized order (Appendix~\ref{app:slotbalance}) and its associated criterion $c$, each judge returns its own preference over the pair.
Its verdict $J_k(p, c) \in \{0, 1\}$ records whether that preference agrees with the constructed synthetic label. Judges have veto power: a score of 1 retains the pair, while 0 discards it. With $\mathcal{D}_0$ representing the initial set of generated pairs and $\mathcal{D}_k$ the subset surviving stage $k$, the cascade progresses as:
\begin{equation}
\mathcal{D}_k = \bigl\{\, p \in \mathcal{D}_{k-1} \ \mid \ J_k(p, c) = 1 \,\bigr\}, \quad k = 1, \dots, K.
\label{eq:cascade}
\end{equation}
This construction forms a nested sequence $\mathcal{D}_K \subseteq \cdots \subseteq \mathcal{D}_1 \subseteq \mathcal{D}_0$, ensuring a pair is preserved only if approved by all $K$ judges.
We set $K = 2$ and evaluate intermediate depths (Table~\ref{tab:arms}).

\paragraph{Retry}
To improve data efficiency, a cell rejected by the filtering cascade is resampled with the same steering prompts and fresh rollouts, up to $R$ retries.
Resampled trajectories reuse the steering instructions to ensure level contrast stratification.
Since retries target cells previously rejected by the filter, subsequent attempts operate on inherently harder tasks, leading to an expected drop in retention rate but raising the overall data collection count.
Thus, $R$ serves as a hyperparameter trading increased dataset yield against compute cost.
We set $R = 2$ in our experiment and analyze this trade-off in Table~\ref{tab:funnel}.

The complete dataset curation pipeline, integrating generation, filtering, and retries, is summarized in Algorithm~\ref{alg:synth} (Appendix~\ref{app:notation}).

\subsection{Meta-Evaluating an Evaluator}
\label{sec:metaeval}

An evaluator under test is a scoring function $f(x; c) \in [0, 1]$ that rates a single trajectory $x$ against a criterion $c$. Because $f$ evaluates each trajectory independently without viewing trajectory pairs, its pairwise preference over $p = (x^-, x^+)$ is derived directly from the score gap:
\begin{equation}
\Delta_f(p) = f(x^{+}; c) - f(x^{-}; c),
\qquad
\hat{y}_f(p) = \operatorname{sign} \Delta_f(p) \in \{+1, -1, 0\},
\label{eq:gap}
\end{equation}
where $\hat{y}_f(p) = 0$ indicates a tie in score. This setup mirrors application scenarios where evaluators score individual execution traces without access to counterfactual rollouts.

Because synthetic ground truth $\star$ always prefers $x^+$, an evaluator aligns with ground truth if and only if it assigns a strictly higher score to $x^+$ ($\Delta_f(p) > 0$).
Ties ($\Delta_f(p) = 0$) count as disagreements because a deployed evaluator that cannot separate $x^{+}$ from $x^{-}$ supplies no usable signal.
Evaluator accuracy is therefore defined as the agreement between the evaluator and the synthetic ground truth:
\begin{equation}
\mathrm{Agr}(f, \star; \mathcal{D}) = \frac{1}{|\mathcal{D}|} \sum_{p \in \mathcal{D}} \mathbf{1}\bigl[\Delta_f(p) > 0\bigr].
\label{eq:eval_acc}
\end{equation}
More broadly, we can formalize any pairwise decision source as a \emph{preference provider} $r$ with decision space $\hat{y}_r(p) \in \{+1, -1, 0\}$. Evaluators may output neutral ties ($\hat{y}_f = 0$), whereas human annotations and synthetic ground truth are strictly binary in $\{-1, +1\}$. Agreement between any two providers $r$ and $r'$ over a dataset $\mathcal{D}$ is:
\begin{equation}
\mathrm{Agr}(r, r'; \mathcal{D}) = \frac{1}{|\mathcal{D}|} \sum_{p \in \mathcal{D}} \mathbf{1}\bigl[\, \hat{y}_r(p) = \hat{y}_{r'}(p) \,\bigr].
\label{eq:agr}
\end{equation}

In the experiments below, we assess evaluator accuracy $\mathrm{Agr}(f, \star; \mathcal{D})$ (Section~\ref{sec:discriminates}), synthetic label accuracy against human majority ($\mathrm{maj}$) annotations $\mathrm{Agr}(\mathrm{maj}, \star; \mathcal{D}_k)$ (Section~\ref{sec:human}), and fine-grained evaluations across criteria, domains, level contrasts, and agent models (Appendices~\ref{app:strata} and~\ref{app:evaluators_all}).

\section{Experimental Setup}
\label{sec:setup}

\paragraph{Agent System \& Criteria}
We build an agent trajectory collection system upon a variant of the $\tau^3$-bench repository \citep{yao2025taubench, barres2026taubench, shi2026tauknowledge}.
We rely on its task definitions, agent framework, user simulator, tool sets, and domain-specific environments, but use our own LM evaluator system.
This setup mimics the expected usage of the PADM\'E algorithm in real-world scenarios, where developers of an agent system bring in the agents, task parameters, and evaluation criteria of interest, and meta-evaluate the evaluator over them with PADM\'E.
The base version of $\tau^3$-bench comprises 375 tasks across four task domains: airline (50), banking (97), retail (114), and telecom (114), each featuring different agent prompts, environments, and tools.
We evaluate three runtime criteria: \emph{friendliness}, \emph{communication clarity}, and \emph{task resolution}.
Detailed descriptions of the criteria are in Appendix~\ref{app:prompt_criteria}.
The task-criterion matrix contains 1,125 evaluation cells, with quality level contrasts stratified evenly across the dataset.

\paragraph{Pipeline Models \& Data Curation}
All pipeline components are driven by open-weight SLMs ranging from 3B to 5.1B active parameters (21B to 117B total parameters).
We test agent trajectory pairs generated using \texttt{gpt-oss-20b}, \texttt{gpt-oss-120b}, and \texttt{nemotron-lightning-3.5}, with both rollouts in a pair generated by the same model.
User interactions are simulated via \texttt{qwen3-30b-a3b-instruct}.
Steering instructions are generated by \texttt{gpt-oss-120b}.
Filtering employs $K=2$ judges ($J_1 = \text{\texttt{nemotron-lightning-3.5}}$ and $J_2 = \text{\texttt{gpt-oss-120b}}$).
Each data point is under a retry budget of $R=2$.
Running the 1,125 initial cells under this setup yields 1,000 kept pairs, distributed over the four dataset axes as Table~\ref{tab:axes} in Appendix~\ref{app:dataset} shows.
For all open-weight models, we use the Fireworks model deployment and serverless access service\footnote{\url{https://fireworks.ai/models}.} (Appendix~\ref{app:licenses}).

\paragraph{Meta-Evaluation Sweep}
We evaluate 25 LMs as evaluators across open-weight and proprietary model families on all 1,000 kept pairs across 3 runs (Appendix~\ref{app:licenses}).
The open-weight models contain 2B to 2.8T parameters.
The evaluator prompt can be found in Appendix~\ref{app:prompt_evaluator}.
Model evaluations run under vendor-default reasoning budgets and at temperature 0, unless temperature cannot be set\footnote{This refers to the more recent OpenAI GPT family of proprietary models.}.
Three of the 25 evaluators overlap with models used in the generation and filtering pipeline and are explicitly flagged in downstream analyses for potential contamination (Appendix~\ref{app:contamination}).

\paragraph{Human Validation Protocol}
Human validation is conducted on a stratified sample of 150 pairs from the \textit{initial generation attempts ($r=0$)} across all criteria and domains (representativeness checks in Appendix~\ref{app:panel_poststrat}).
It serves exclusively to validate the data curation pipeline rather than operating as part of the pipeline itself.
Six human annotators form two panels of three on disjoint sets of 75 pairs, generating 450 binary preference choices along with confidence ratings.
The majority vote across the three annotators serves as the human reference label ($\mathrm{maj}$).
Additional information can be found in Appendix~\ref{app:panel}.

\section{Results and Analysis}
\label{sec:results}

\subsection{Dataset}
\label{sec:produced}

Out of 1,125 initial task-criterion cells, 1,000 trajectory pairs survive both filter stages under a retry budget of $R=2$, requiring 1,673 total generation draws (Table~\ref{tab:funnel}).
Per-attempt retention decays predictably across retries, as subsequent retries operate exclusively on previously rejected cells.
Retries yield an additional 230 validated pairs (+30\%) at the cost of 548 supplementary trajectory rollouts.
Across all subsets, the first filter judge $J_1$ accounts for the vast majority of rejections. Detailed subset breakdowns are provided in Appendix~\ref{app:yield}.

The dataset curation pipeline generates 1,000 validated trajectory pairs at a total cost of \$23.63, or \$0.024 per kept pair. Details of cost are in Appendix~\ref{app:cost}.

\begin{table}[H]
\caption{\textbf{Data curation funnel over 1,125 cells across retry attempts.}}
\label{tab:funnel}
\centering
\scriptsize
\begin{tabular}{@{}lrrrrrr@{}}
\toprule
retry (r) & draws & kept & $J_1$ rejected & $J_2$ rejected & retention rate (\%) & cumulative retention rate (\%) \\
\midrule
$0$   & 1{,}125 & 770   & 266 & 89  & 68.4 & 68.4 \\
$1$   & 355     & 161   & 146 & 48  & 45.4 & 82.8 \\
$2$   & 193     & 69    & 100 & 24  & 35.8 & 88.9 \\
\midrule
total                   & 1{,}673 & 1{,}000 & 512 & 161 & 59.8 & 88.9 \\
\bottomrule
\end{tabular}
\end{table}

\subsection{Human Study}
\label{sec:human}

\paragraph{Annotator Confidence and Dataset Difficulty}
Annotators report their confidence on each pair as 0 (a guess), 1 (leaning), or 2 (certain).
We use the per-pair mean over the three annotators as a proxy for how difficult that pair is to judge.
Mean self-reported annotator confidence increases by $+0.02$ on the 0 to 2 scale (1\%) across filtering stages, though the influence of $J_1$ and $J_2$ differs (Appendix~\ref{app:panel_confidence}).
This suggests that filtering does not significantly trivialize the resulting data.

\paragraph{Human Alignment with Labels}
Filtering monotonically increases alignment between the synthetic label ($\star$) and human majority vote ($\mathrm{maj}$), as computed by synthetic label accuracy $\mathrm{Agr}(\mathrm{maj}, \star; \mathcal{D}_k)$ with Eq.~\eqref{eq:agr}.
Passing pairs through the filters boosts label validity from 73.3\% to 84.6\% while shifting human-label agreement (Krippendorff's $\alpha$) from $+0.468$ to $+0.694$ (Table~\ref{tab:arms}; panel reliability in Appendix~\ref{app:panel_alpha}).

The improvement appears to plateau at $K = 2$.
On the subset of pairs rejected by $J_1$ ($\mathcal{D}_0 \setminus \mathcal{D}_1$), label validity falls to 38.9\% (14/36), which confirms that the filter selectively removes misaligned or noisy trajectories.
Label validity among $J_2$-rejected pairs is 80.0\% (8/10), close to the 84.6\% among retained pairs.
Detailed subset breakdowns are provided in Appendix~\ref{app:strata}.

\begin{table}%
\caption{
\textbf{Agreement between synthetic labels and human majority vote across filter depths $\mathbf{\mathcal{D}_k}$.}
Yield indicates the proportion of the 150 annotated pairs surviving at each depth.
Synthetic label accuracy reports $\mathrm{Agr}(\mathrm{maj}, \star; \mathcal{D}_k)$ with bootstrap 95\% confidence intervals in brackets.
$\alpha$ is Krippendorff's $\alpha$ for the same label-versus-majority decision.}
\label{tab:arms}
\centering
\scriptsize
\begin{tabular}{@{}lrrlr@{}}
\toprule
dataset & $|\mathcal{D}|$ & yield & synthetic label accuracy & $\alpha$ \\
\midrule
$\mathcal{D}_0$ (unfiltered) & 150 & 100\% & 73.3\% [66--80] & +0.468 \\
$\mathcal{D}_1$ ($J_1$)     & 114 & 76\%  & 84.2\% [76--90] & +0.686 \\
$\mathcal{D}_2$ ($J_1\&J_2$)     & 104 & 69\%  & \textbf{84.6\%} [76--90] & \textbf{+0.694}\\
\bottomrule
\end{tabular}
\end{table}

\subsection{Meta-Evaluation Sweep}
\label{sec:discriminates}

To establish a performance spectrum across common language models, we benchmark 25 LMs with the same basic evaluator prompt (Appendix~\ref{app:prompt_evaluator}).
We have each model score all 2{,}000 trajectories (1{,}000 pairs) independently across three separate runs.
Table~\ref{tab:evaluators} lists 12 model results out of 25 for readability.
Two numbers directly inform the performance of each model: \textbf{accuracy} (Equation~\ref{eq:eval_acc}), our most important performance metric, and \textbf{average score standard deviation (SD)}\footnote{Note that average score SD is not the standard deviation of all scores assigned by each evaluator across the dataset, but the average of the standard deviation of each data point score across 3 runs for each evaluator.}, which helps us understand the stability of the evaluator's scoring.
All analyses in this section are conducted over the full 25 model results, which are reported in Table~\ref{tab:evaluators_full} in the appendix.

\begin{table}[H]
\caption{
\textbf{Performance of evaluators across 1{,}000 trajectory pairs ($n = 3$ runs).}
12 of 25 evaluators are selected to span the accuracy range and model families (full view in Table~\ref{tab:evaluators_full}). All reported statistics are computed across all 25 evaluators. Released indicates public release month. Temperature = 0 except where unavailable$^{\ddagger}$. Preferences derive from individual trajectory score gaps; ties count as incorrect. Accuracy and score denote benchmark level accuracy and raw evaluator score, respectively. SD is standard deviation. Leniency is mean emitted score. Average score SD measures per data point scoring stability across 3 runs. \textbf{Bold} indicates best performance per column across all 25 evaluators. Accuracy and average score SD serve as the two primary evaluator metrics.}
\label{tab:evaluators}
\centering
\footnotesize
\setlength{\tabcolsep}{1.5pt}
\resizebox{\textwidth}{!}{%
\begin{tabular}{@{}rlrrrccrrrrrrrrrrr@{}}
\toprule
& & & \multicolumn{2}{c}{parameters} & & accuracy (\%) & leniency & average & tie & \multicolumn{4}{c}{by domain (\%)} & \multicolumn{3}{c}{by criterion (\%)} & \# distinct \\
\cmidrule(lr){4-5}\cmidrule(lr){11-14}\cmidrule(lr){15-17}
\# & evaluator & released & total & active & reas. & mean $\pm$ SD & (mean score) & score SD & (\%) & airl. & bank. & retail & telec. & clar. & friend. & task & scores \\
\midrule
1 & \texttt{claude-opus-5} & 2026-07 & closed & closed & Yes & \textbf{85.1} $\pm$ 0.93 & 0.324 & 0.023 & 4.1 & \textbf{84} & \textbf{82} & \textbf{86} & \textbf{87} & 73 & 96 & \textbf{85} & 69 \\
3 & \texttt{kimi-k3} & 2026-07 & 2.8T & 104B & Yes & 83.6 $\pm$ 0.12 & 0.508 & 0.036 & 6.0 & 83 & 80 & 84 & \textbf{87} & 71 & \textbf{97} & 81 & 37 \\
6 & \texttt{glm-5p2} & 2026-06 & 753B & $\sim$30B & Yes & 81.5 $\pm$ 0.50 & 0.521 & 0.051 & 10.1 & \textbf{84} & 81 & 81 & 82 & \textbf{78} & 92 & 73 & 31 \\
8 & \texttt{gpt-5.4-nano}$^{\ddagger}$ & 2026-03 & closed & closed & No & 79.4 $\pm$ 1.16 & 0.631 & 0.052 & 7.8 & 81 & 78 & 78 & 81 & 74 & 84 & 80 & 61 \\
10 & \texttt{gpt-5.6-sol}$^{\ddagger}$ & 2026-07 & closed & closed & Yes & 78.7 $\pm$ 0.40 & 0.543 & 0.035 & 4.9 & 79 & 77 & 81 & 77 & 64 & \textbf{97} & 74 & 86 \\
11 & \texttt{claude-haiku-4-5} & 2025-10 & closed & closed & No & 78.7 $\pm$ \textbf{0.10} & 0.505 & \textbf{0.000} & 11.6 & 81 & 79 & 76 & 81 & 76 & 85 & 74 & 32 \\
13 & \texttt{nemotron-3-ultra} & 2026-06 & 549B & 55B & Yes & 76.8 $\pm$ 0.76 & 0.558 & 0.044 & 14.2 & 75 & 80 & 70 & 81 & 71 & 88 & 71 & 28 \\
14 & \texttt{llama3.1-70b} & 2024-07 & 70.6B & dense & No & 70.0 $\pm$ 0.55 & 0.632 & 0.016 & 22.8 & 67 & 73 & 67 & 72 & 64 & 69 & 77 & 14 \\
15 & \texttt{gemini-3.7-flash} & 2026-08 & closed & closed & Yes & 69.5 $\pm$ 0.45 & 0.615 & 0.026 & 23.4 & 71 & 70 & 69 & 69 & 57 & 95 & 54 & 34 \\
17 & \texttt{gemma-4-31b} & 2026-03 & 32.2B & dense & Yes & 68.1 $\pm$ 0.15 & 0.657 & 0.023 & 27.4 & 63 & 67 & 66 & 74 & 47 & 92 & 63 & 14 \\
18 & \texttt{deepseek-v4-pro} & 2026-08 & 1.6T & 49B & Yes & 67.0 $\pm$ 0.67 & 0.505 & 0.045 & 20.4 & 62 & 70 & 69 & 65 & 61 & 90 & 49 & 23 \\
25 & \texttt{qwen3-4b} & 2025-08 & 4.4B & dense & No & 36.8 $\pm$ 0.65 & 0.804 & 0.029 & 56.9 & 36 & 34 & 32 & 44 & 30 & 25 & 55 & 17 \\
\midrule
\multicolumn{6}{@{}l}{\emph{Mean, all 25 evaluators}} & 70.1 &  &  &  & 68.7 & 69.9 & 69.0 & 72.2 & 62.0 & 79.5 & 67.9 &  \\
\bottomrule
\end{tabular}}
\end{table}

\paragraph{Subset Analysis}
We discuss four subsets: evaluation criteria (friendliness, communication clarity, task resolution), domain (retail, telecom, banking, airline), level contrast (\textsf{bad-ok}, \textsf{ok-good}, \textsf{bad-good}), and agent model (\texttt{nemotron-lightning-3.5}, \texttt{gpt-oss-120b}, \texttt{gpt-oss-20b}).

The \textsf{bad}--\textsf{ok} contrast (73.2\%) and the \textsf{bad}--\textsf{good} contrast (73.6\%) have similar accuracy, while the accuracy of the \textsf{ok}--\textsf{good} subset is much lower, at 63.4\% (Table~\ref{tab:evaluators_full}).
A hypothesis is that, compared to detecting bad trajectories, it is harder for evaluators to distinguish the relative performance of two acceptable trajectories.

\texttt{gpt-oss-120b}'s trajectories have an average accuracy of 74.9\%, \texttt{gpt-oss-20b} 70.5\%, and \texttt{nemotron-lightning-3.5} 65.1\% (Table~\ref{tab:evaluators_full}).
This suggests that the model backbone of the agent impacts the difficulty of evaluating the agent trajectories.

\paragraph{Correlation Analysis}
Table~\ref{tab:correlations} shows the correlation analyses between accuracy and seven factors.

Accuracy is highly correlated with scoring resolution.
Tie rates range from 3.4\% to 56.9\% and correlate strongly with accuracy (Spearman's $\rho = -0.946$).
In many cases, models fail due to coarse scoring granularity rather than misjudging.
Across models, distinct score count correlates with accuracy at $\rho = +0.797$, showing that score resolution is important for discriminative ability.

\begin{table}%
\caption{\textbf{Seven evaluator properties against accuracy, sorted by Spearman's correlation coefficients.}
$p_{\mathrm{holm}}$ is the Holm-Bonferroni corrected p-value across the seven tests, and \textbf{bolded} where it indicates significance.
$n = 15$ on some rows because ten closed models do not publish parameter counts.
Reasoning is a binary factor.}
\label{tab:correlations}
\centering
\scriptsize
\newcolumntype{F}{>{\raggedright\arraybackslash}p{2.60cm}}
\newcolumntype{N}{>{\raggedleft\arraybackslash}p{0.40cm}}
\newcolumntype{C}{>{\raggedleft\arraybackslash}p{0.90cm}}
\newcolumntype{P}{>{\raggedleft\arraybackslash}p{1.20cm}}
\begin{tabular}[t]{@{}FNCP@{}}
\toprule
factor & $n$ & $\rho$ & $p_{\mathrm{holm}}$ \\
\midrule
tie rate                  & 25 & $-0.946$ & $\mathbf{<0.0001}$ \\
distinct score values     & 25 & $+0.797$ & $\mathbf{<0.0001}$ \\
leniency (mean score)     & 25 & $-0.725$ & $\mathbf{0.0002}$ \\
total parameters          & 15 & $+0.725$ & $\mathbf{0.0089}$ \\
\bottomrule
\end{tabular}
\hfill
\begin{tabular}[t]{@{}FNCP@{}}
\toprule
factor & $n$ & $\rho$ & $p_{\mathrm{holm}}$ \\
\midrule
active parameters         & 15 & $+0.518$ & 0.1347 \\
reasons by default        & 25 & $+0.404$ & 0.1347 \\
release date             & 25 & $+0.359$ & 0.1347 \\
                          &    &          &        \\
\bottomrule
\end{tabular}
\end{table}

The mean score an evaluator assigns to data points, which is a way of quantifying leniency, shows a significant anticorrelation with the evaluator performance ($\rho = -0.725$). It is possible that a harsher evaluator holds a longer internal list of expectations of the agent behavior, and is thus better at distinguishing nuanced differences between two trajectories. An evaluator that is easily satisfied, i.e., assigns scores close to 1 easily, risks conflating good and great agent behaviors.

Total parameter ($\rho = +0.725$) and active parameter counts ($\rho = +0.518$, not significant) show some level of correlation with accuracy. For proprietary models, an exception is that \texttt{gpt-5.4-nano} and \texttt{gpt-5.4-mini}, which have smaller expected parameter counts, outperform \texttt{gpt-5.6-sol}.

\section{Limitations and Future Work}
\label{sec:limitations}

\textbf{Extensibility to More Evaluation Criteria:}
Future work should apply PADM\'E to a larger and more diverse set of criteria.
It would especially benefit from a stress test of criteria that, through steering in a direction, would go against critical instruction-following training of the agent models (such as toxicity level or answer safety).
\textbf{Broadening Agent and Domain Coverage:}
The current study limits agent trajectory generation to $\tau^3$-bench and its domains.
To further validate the proposed method, it should be generalized across diverse agent architectures, operation environments, and broader domain benchmarks beyond customer-service agent operations.
\textbf{Score Granularity and Calibration:}
Given that evaluator accuracy is heavily driven by score resolution, future work should systematically evaluate controlled scoring regimes (e.g., discrete Likert scales versus continuous $[0, 1]$ bounds constrained to fixed decimal precision) to explore how output formatting affects model ties and discrimination.
\textbf{Other Reliability Signals:}
For criteria with programmatic rewards, reliability can be assessed without human annotations.
We briefly discuss the alignment between verifiable rewards with synthetic labels in Appendix~\ref{sec:env_reward}, but it warrants further investigation.

\section{Conclusion}
\label{sec:conclusion}

Meta-evaluation is the evaluation of LM evaluator reliability.
Adopting preference alignment in place of absolute score alignment, we demonstrate that meta-evaluation data can be synthesized rather than manually annotated.
The proposed framework, PADM\'E, steers an agent's trajectories along criteria axes, uses the steering intents as the initial labels, and applies peer-sized judges to veto trajectory contrasts.
Using models at or below 5.1B active parameters, PADM\'E constructs a 1,000-pair benchmark across four domains and three criteria at a low cost.
Filtering boosts human-label agreement from 73.3\% to 84.6\%.
The resulting dataset separates 25 candidate evaluators across a 48.3-percentage-point accuracy spread.
The data can be regenerated whenever criteria are modified or agents are updated without human intervention.
We thus present a generalizable data synthesis and meta-evaluation recipe rather than a static benchmark.

\begin{ack}

All authors are employees at Uniphore. All funding is provided by Uniphore.

The authors would like to thank Ishika Agarwal, Bowen He, Tommy Li, Ethan Soon, Artin Tajdini, and Ming Xin (ordered by last names) for their help as human annotators.
\end{ack}

{
\small
\bibliographystyle{unsrtnat}
\bibliography{refs}
}

\newpage
\appendix

\input{appendix.tex}

\end{document}

%% file: appendix.tex
\section{Notations and the Algorithm}
\label{app:notation}

\begin{table}[H]
\caption{\textbf{Notation, grouped by pipeline stage in the order the paper uses it.}}
\label{tab:notation}
\centering
\small
\begin{tabular}{@{}l p{0.70\textwidth}@{}}
\toprule
\multicolumn{2}{@{}l}{\emph{Generation}} \\
$t$ & a task from the dataset, carrying its domain policy, tools, and user goal. \\
$c$ & a criterion, supplied at runtime as free text: a name and a detailed description. \\
$\ell \in \mathcal{L}$ & a steer level, $\mathcal{L} = \{\textsf{bad} \prec \textsf{ok} \prec \textsf{good}\}$, ordered by intended quality. \\
$\theta$ & everything held fixed while the steer level varies: agent model, user simulator, domain, decoding seed. \\
$s_{c,t,\ell}$ & the steering instruction, written by a small model from $c$, $t$, and $\ell$, Eq.~\eqref{eq:gen}. \\
$x$ & a trajectory: the complete trace generated by an interactive dialogue between a user simulator and an agent executing a task, with tool calls and its outcome. \\
$p = (x^-, x^+)$ & a pair, where $x^+$ is generated at the higher steer level. Slot order is randomized whenever the pair is shown to an LLM filter or a human annotator. \\
\midrule
\multicolumn{2}{@{}l}{\emph{Curation}} \\
$K$ & the number of filter judges applied in sequence. \\
$R$ & the retry budget: a cell may be drawn up to $1 + R$ times. \\
$J_k$ & the $k$-th filter judge, $J_k(p,c) \in \{0, 1\}$. \\
$\mathcal{D}_k$ & the pairs surviving the first $k$ judges, $0 \le k \le K$, Eq.~\eqref{eq:cascade}; $\mathcal{D}_0$ is every drawn pair. \\
\midrule
\multicolumn{2}{@{}l}{\emph{Evaluation}} \\
$f$ & the evaluator under test, $f(x; c) \in [0, 1]$, applied to one trajectory at a time. \\
$\Delta_f(p)$ & its score gap on a pair, Eq.~\eqref{eq:gap}. \\
$r$ & a preference provider: any source of a pairwise decision. Instances used here are the evaluator under test $f$ (preference derived from $\Delta_f(p)$), a filter judge $J_k$, a human annotator $a$, the panel majority $\mathrm{maj}$ of three annotators, and the synthetic label $\star$. \\
$\hat{y}_r(p)$ & the preference of provider $r$ on pair $p$: $+1$ for $x^+$, $-1$ for $x^-$, $0$ for none. \\
$\star$ & the constructed synthetic label, treated as a preference provider: $\hat{y}_\star \equiv +1$ by definition. \\
\midrule
\multicolumn{2}{@{}l}{\emph{Reporting}} \\
$\mathrm{Agr}(r, r'; \mathcal{D})$ & how often two preference providers agree over $\mathcal{D}$, Eq.~\eqref{eq:agr}. \\
$\mathcal{D}|_z$ & $\mathcal{D}$ restricted to a subset $z$: one criterion, domain, level contrast, or agent. \\
\bottomrule
\end{tabular}
\end{table}

Table~\ref{tab:notation} collects the notation used in the main paper, grouped by pipeline stage.

\begin{algorithm}[H]
\caption{\textbf{PADM\'E data curation algorithm.} Each cell yields at most one kept pair.}
\label{alg:synth}
\begin{algorithmic}[1]
\Require cells $\{(t,c)\}$, fixed context $\theta$, judges $J_1, \dots, J_K$, retry budget $R$
\State $\mathcal{D}_k \gets \emptyset$ for $k = 0, \dots, K$
\ForAll{cells $(t, c)$}
  \State take the level contrast $\ell \prec \ell'$ for this cell from the stratified schedule
  \For{$\text{attempt} = 1$ \textbf{to} $1 + R$}
    \State $s_{c,t,\ell} \gets \textsc{WriteSteer}(c, t, \ell)$, \quad $s_{c,t,\ell'} \gets \textsc{WriteSteer}(c, t, \ell')$
    \State $x^{-} \sim \mathrm{Roll}(t,\, s_{c,t,\ell};\, \theta)$,
           \quad $x^{+} \sim \mathrm{Roll}(t,\, s_{c,t,\ell'};\, \theta)$
    \State $p \gets (x^{-}, x^{+})$, \quad
           $\mathcal{D}_0 \gets \mathcal{D}_0 \cup \{p\}$
    \State $k \gets 0$
    \While{$k < K$ \textbf{and} $J_{k+1}(p,c) = 1$}
      \State $k \gets k + 1$, \quad $\mathcal{D}_k \gets \mathcal{D}_k \cup \{p\}$
      \Comment{a veto stops the cascade}
    \EndWhile
    \If{$k = K$}
      \State \textbf{break}
      \Comment{cell filled; only a rejected cell is ever redrawn}
    \EndIf
  \EndFor
\EndFor
\State \Return $\mathcal{D}_K$
  \Comment{$\mathcal{D}_0, \dots, \mathcal{D}_{K-1}$ are kept for the filter-depth ablation}
\end{algorithmic}
\end{algorithm}

The complete PADM\'E dataset curation pipeline, integrating generation, filtering, and retries, is summarized in Algorithm~\ref{alg:synth}.

\section{Dataset Composition and Evaluation Protocols}
\label{app:dataset}
Table~\ref{tab:axes} outlines the composition of the 1{,}000-pair shipping dataset across its four principal axes.
Table~\ref{tab:roles} details the input constraints, task formulations, and output formats for each preference provider.

The pipeline naturally produces a balanced dataset across criteria, domains, level contrasts, and agent models without requiring artificial quotas or post-hoc rebalancing, maintaining strong yield consistency across subsets (detailed further in Table~\ref{tab:retention} and Appendix~\ref{app:yield}).

The filter judge ($J_k$) and the human annotator ($a$) operate under identical information availability.
Both receive identical contextual fields, perform pairwise trajectory comparisons, and remain strictly blinded to steering metadata.
This symmetry ensures that human label alignment serves as a direct validation of the cascade filter rather than an artifact of information asymmetry. In contrast, the evaluator under test ($f$) operates pointwise on single trajectories without cross-trajectory visibility, which mirrors realistic deployment conditions.

\begin{table}[H]
\caption{\textbf{A glossary of dataset subsets, along with the distribution of the 1{,}000 pairs across axes.} Each pair carries exactly one level per axis, resulting in subtotal sums of 1{,}000 per block.
Comprehensive retention rates per subset appear in Table~\ref{tab:retention}. Evaluation criterion descriptions reflect the full runtime definitions, also provided in Appendix~\ref{app:prompt_criteria}.
Stratified experimental breakdowns are reported in Appendix~\ref{app:strata}.}
\label{tab:axes}
\centering
\small
\begin{tabular}{@{}lrp{0.6\textwidth}@{}}
\toprule
level & pairs & description \\
\midrule
\multicolumn{3}{@{}l}{\emph{Evaluation criterion}} \\
\quad friendliness          & 355 & Warmth and consideration toward the customer: whether the agent acknowledges their situation and how they feel about it, delivers unwelcome news with care, and leaves them feeling attended to. Judge the manner, not whether the request was resolved. \\
\quad task resolution       & 329 & Whether the customer's actual problem was settled: did the agent establish what was needed, take the actions that would resolve it, and leave the customer with the outcome they came for. A correct refusal counts as resolution -- if the request was not permitted, saying so plainly and explaining why resolves it, while quietly doing it anyway does not. Judge the outcome, not the manner or how well it was explained. \\
\quad communication clarity & 316 & How easily the customer can follow the agent: whether the main point is findable, whether technical or policy language is explained, whether multi-part information is organised, and whether the customer is left knowing what is true and what happens next. Judge the presentation, not the warmth or the outcome. \\
\midrule
\multicolumn{3}{@{}l}{\emph{Domain}} \\
\quad retail  & 309 & 114 tasks; 1{,}158-word policy, 16 agent tools \\
\quad telecom & 294 & 114 tasks; 3{,}715-word policy, 13 agent tools, 30 user-side tools \\
\quad banking & 262 & 97 tasks; 926-word policy, 16 agent tools, including retrieval \\
\quad airline & 135 & 50 tasks; 1{,}313-word policy, 14 agent tools \\
\midrule
\multicolumn{3}{@{}l}{\emph{Level contrast}} \\
\quad \textsf{bad}--\textsf{good} & 359 & wider gap \\
\quad \textsf{ok}--\textsf{good}  & 328 & narrower gap \\
\quad \textsf{bad}--\textsf{ok}   & 313 & narrower gap \\
\midrule
\multicolumn{3}{@{}l}{\emph{Agent model}} \\
\quad \texttt{nemotron-lightning-3.5} & 341 & 32B total, 3B active \\
\quad \texttt{gpt-oss-120b}               & 334 & 116.8B total, 5.1B active \\
\quad \texttt{gpt-oss-20b}                & 325 & 20.9B total, 3.6B active \\
\bottomrule
\end{tabular}
\end{table}

\begin{table}%
\caption{\textbf{A glossary of information exposure, task objectives, and output specifications across preference providers.} Symmetrical blinding ensures human annotators directly benchmark filter judge decisions, whereas evaluators under test operate in single-trajectory pointwise mode to match real-world deployment.}
\label{tab:roles}
\centering
\small
\begin{tabular}{@{}p{0.115\textwidth}p{0.26\textwidth}p{0.25\textwidth}p{0.26\textwidth}@{}}
\toprule
 & filter judge $J_k$ & human annotator $a$ & evaluator under test $f$ \\
\midrule
purpose      & curates the dataset & validates the dataset & the object of measurement \\
\midrule
is given     & the criterion name and definition, the agent's tools and the domain policy, and both trajectories & the same five fields, rendered in a purpose-built interface (Appendix~\ref{app:panel}) & the criterion name and definition, the agent's tools and the domain policy, and one trajectory \\
\midrule
is not given & the steering prompts, the target levels, or which side was steered better & the same, and no filter verdict & the same, and never the other trajectory \\
\midrule
is asked     & which of the two is better on the criterion & which of the two is better on the criterion & to score the agent on the criterion from $0.0$ to $1.0$ \\
\midrule
slot order   & randomized per pair and per judge & randomized per pair & no order arises: one trajectory per call \\
\midrule
returns      & a forced choice of one side, with at most three sentences of reasoning & a forced choice of one side, with a confidence rating of $\{0,1,2\}$ & a score in $[0, 1]$, with at most three sentences of reasoning \\
\midrule
preference   & the side it chose; the pair survives only when that matches the synthetic label & the side it chose; the majority vote of three is $\mathrm{maj}$ & the sign of the score gap (Eq.~\eqref{eq:gap}) \\
\bottomrule
\end{tabular}
\end{table}

\section{Dataset Audits and Yield Dynamics}
\label{app:generation}

\subsection{Generation Yield and Retention}
\label{app:yield}

Generation retention is reported per experimental subset rather than artificially constrained. Because no post-hoc rebalancing is applied to force quota targets, Table~\ref{tab:retention} directly reflects where trajectory pairs easily satisfy judge filtering versus where narrower quality gaps require higher generation compute and retries.

Retention yield strictly tracks two primary factors: the target level contrast and the specific evaluation criterion. First, wider level contrasts yield higher retention rates: \textsf{bad}--\textsf{good} contrasts complete 96\% of targeted cells at 1.38 draws per retained pair, compared to 84\% completion at 1.90 draws per pair for narrower \textsf{bad}--\textsf{ok} contrasts. Second, criteria evaluating overt linguistic style achieve higher retention than structural task completion metrics: \texttt{friendliness} reaches 95\% cell retention, outperforming \texttt{task\_resolution} (88\%) and \texttt{communication\_clarity} (84\%). Across domains, retention rates remain tightly clustered within a 4-percentage-point band, with telecom recording the lowest completion rate (86\%) due to its underlying tool and policy complexity (Table~\ref{tab:domains}).

\begin{table}%
\caption{\textbf{Cell retention metrics by subset after 2 retries $(R=2)$.} The retention rate measures the proportion of retained final pairs against the total number of generation draws across all attempts. The cumulative retention rate denotes the proportion of retained final pairs relative to the original set of target cells prior to retries. Draws per retention measures the average number of generation attempts required to yield one retained pair.}
\label{tab:retention}
\centering
\small
\begin{tabular}{@{}llrrrr@{}}
\toprule
axis & subset & \shortstack[r]{cells\\filled} & \shortstack[r]{retention\\rate} & \shortstack[r]{cum. retention\\rate} & \shortstack[r]{draws per\\retention} \\
\midrule
\multicolumn{2}{@{}l}{\emph{By criterion}} \\
& friendliness & 355/375 & 70.3\% & 95\% & 1.42$\times$ \\
& task resolution & 329/375 & 58.9\% & 88\% & 1.70$\times$ \\
& communication clarity & 316/375 & 51.9\% & 84\% & 1.93$\times$ \\
\midrule
\multicolumn{2}{@{}l}{\emph{By domain}} \\
& airline & 135/150 & 61.4\% & 90\% & 1.63$\times$ \\
& banking & 262/291 & 61.4\% & 90\% & 1.63$\times$ \\
& retail & 309/342 & 63.1\% & 90\% & 1.59$\times$ \\
& telecom & 294/342 & 54.9\% & 86\% & 1.82$\times$ \\
\midrule
\multicolumn{2}{@{}l}{\emph{By level contrast}} \\
& bad--good & 359/375 & 72.4\% & 96\% & 1.38$\times$ \\
& ok--good & 328/375 & 56.5\% & 88\% & 1.77$\times$ \\
& bad--ok & 313/375 & 52.5\% & 84\% & 1.90$\times$ \\
\midrule
\multicolumn{2}{@{}l}{\emph{By agent model}} \\
& \texttt{gpt-oss-120b} & 334/367 & 66.9\% & 91\% & 1.49$\times$ \\
& \texttt{nemotron-lightning-3.5} & 341/384 & 57.3\% & 89\% & 1.74$\times$ \\
& \texttt{gpt-oss-20b} & 325/374 & 56.1\% & 87\% & 1.78$\times$ \\
\bottomrule
\end{tabular}
\end{table}

\subsection{Position Balance Audit}
\label{app:slotbalance}

To counter positional bias \citep{positionbias}, the presentation order of trajectories in each pair is randomized during meta-evaluation (Table~\ref{tab:roles}). Position balance serves as a diagnostic audit to verify that this randomization prevents positional confounds in filtering and human annotation. As detailed in Table~\ref{tab:slotbalance}, the target trajectory appears in slot two in 511 of the 1{,}000 final pairs (51.1\%), with all experimental subsets remaining within a 45.5--54.3\% range and displaying no significant departure from uniform parity.

\begin{table}%
\caption{
\textbf{Position balance audit of the target (better-steered) trajectory across data subsets.} Across all groups, slot distribution remains statistically indistinguishable from uniform parity (binomial $p > 0.10$), confirming that position bias is not a baseline confounder.
}
\label{tab:slotbalance}
\centering
\small
\begin{tabular}{@{}lrr@{}}
\toprule
axis & slot-2 share range & binomial $p$ range \\
\midrule
criterion       & 48.7--53.8\% & 0.17--0.91 \\
domain          & 46.9--54.1\% & 0.27--0.73 \\
agent model     & 45.5--54.3\% & 0.11--0.22 \\
level contrast  & 49.2--54.0\% & 0.14--0.82 \\
\midrule
whole dataset & 511/1{,}000 = 51.1\% & \\
\bottomrule
\end{tabular}
\end{table}

\subsection{Environment Reward Verification}
\label{sec:env_reward}

Table~\ref{tab:reward} evaluates dataset pairs against $\tau^3$-bench programmatic task success rewards $[0, 1]$, which track database transitions, disclosure verification, and action matching independently of model annotation. Omitting the LLM judge component renders this audit entirely deterministic. Under this setup, steering should produce a strong environment reward shift for direct task execution (\texttt{task\_resolution}) and a weak or non-significant shift for auxiliary stylistic criteria (\texttt{communication\_clarity} and \texttt{friendliness}).

The empirical results confirm this progression. Steering for \texttt{task\_resolution} produces the largest reward increase ($+0.122$, $p < 0.001$), whereas \texttt{friendliness} exhibits no significant shift ($+0.017$, $p = 0.488$), demonstrating clear orthogonality to task outcome. \texttt{communication\_clarity} occupies an intermediate position ($+0.082$, $p < 0.001$), reflecting a real-world dependency where unclear communication occasionally impedes execution workflows. Since the reward shift is smaller than that of explicit task resolution steering, communication clarity remains a distinct behavioral dimension.
Overall, the vast majority of retained pairs are not confounded by outcome variation. Among the 1{,}000 pairs evaluated on both trajectories, 836 (84\%) yield identical environment rewards, with the higher-steered trajectory scoring higher in 11.8\% of pairs and lower in 4.6\%.

\begin{table}%
\caption{\textbf{Environment reward verification comparing deterministic $\tau^3$-bench task success reward gaps across evaluation criteria.} The reward gap represents the score of the higher-steered trajectory minus that of the lower-steered trajectory. Among the 1{,}000 pairs evaluated on both trajectories, 836 pairs (84\%) exhibit identical environment rewards.}
\label{tab:reward}
\centering
\small
\begin{tabular}{@{}lrrrr@{}}
\toprule
criterion & pairs & mean reward gap & better / worse & sign test $p$ \\
\midrule
task resolution       & 329 & $+0.122$ & 49 / 9  & $<0.001$ \\
communication clarity & 316 & $+0.082$ & 40 / 14 & $<0.001$ \\
friendliness          & 355 & $+0.017$ & 29 / 23 & 0.488 \\
\bottomrule
\end{tabular}
\end{table}

\section{Human Panel Reliability and Sampling Robustness}
\label{app:panel}

\subsection{The Annotation Task}
\label{app:panel_task}

\begin{figure}[H]
\centering
\includegraphics[width=\textwidth]{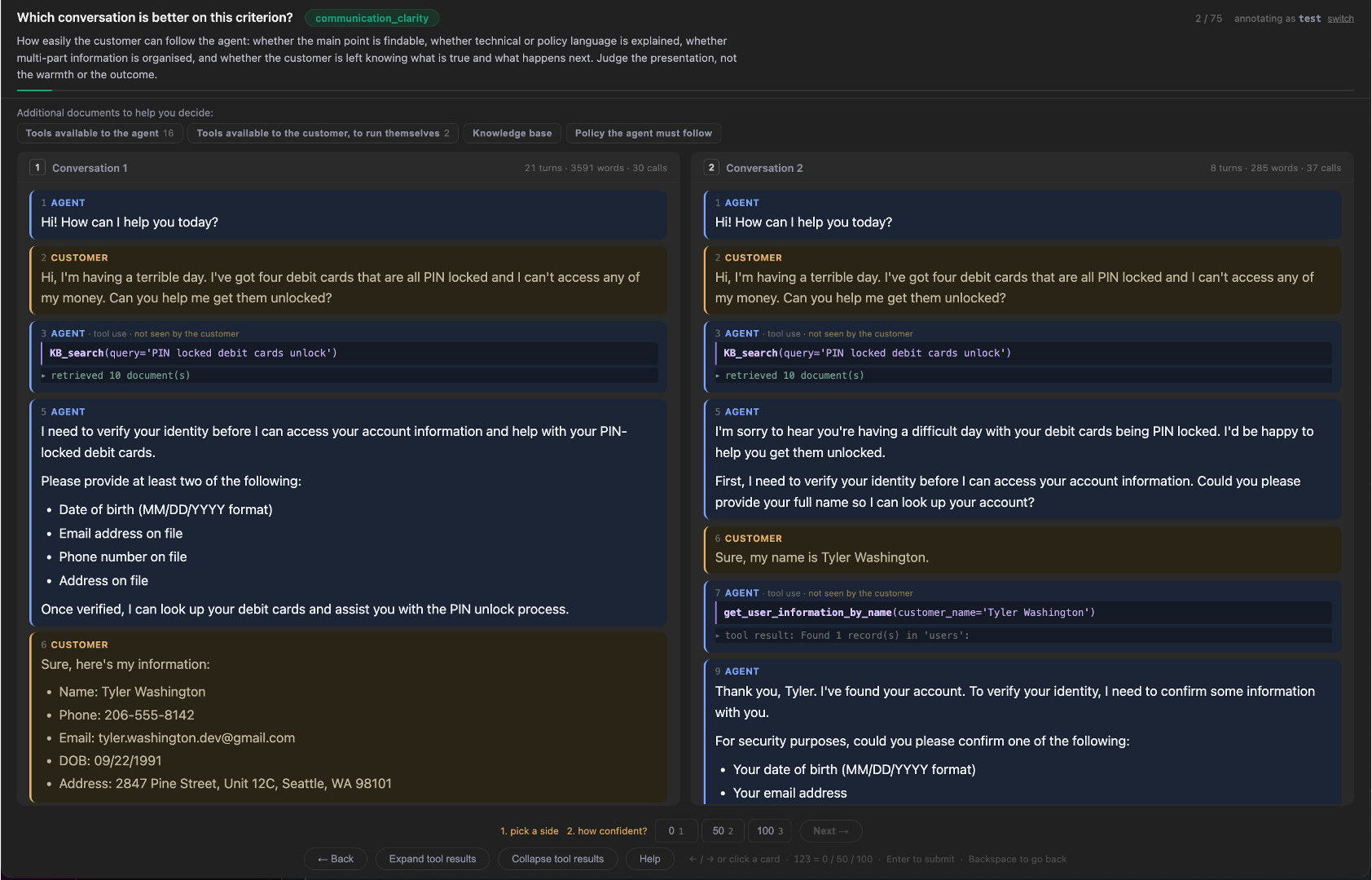}
\caption{\textbf{The interface used by the human panel.} The two trajectories of a pair are presented together with the
criterion being judged; the steering instruction and target levels are withheld, and the annotator returns a
forced binary choice together with a confidence rating.}
\label{fig:ui}
\end{figure}

\paragraph{Annotator Panel}
The panel comprises 3 software engineers and 3 computer science researchers with experience developing language model agents or using them for software development.
Two are full-time employees and four are interns, all within the same company as the authors.
Two are native English speakers and four have professional English fluency.
Annotators review criteria definitions and guidelines prior to evaluation.
The panel is partitioned into two independent groups of three annotators.
Each group evaluates the same 75 trajectory pairs.
Each annotator receives a \$50 stipend for completing the annotations.
The annotation tasks involved evaluating benign model outputs and contained no offensive, sensitive, or harmful material. To minimize psychological fatigue and potential distress, participation was entirely voluntary, annotators were allowed to opt out or take breaks at any time without penalty. No personally identifiable information (PII) was collected.

\paragraph{Task Workflow} For each pair $p = (x^-, x^+)$, annotators select the superior trajectory given the target criterion $c$. Choice is forced ($\hat{y}_a(p) \in \{-1, +1\}$), yielding the majority vote reference $\mathrm{maj}$ used in Section~\ref{sec:human}. Annotators also record confidence on a three-point scale ($0$ for guess, $1$ for leaning, $2$ for certain), analyzed in Appendix~\ref{app:panel_confidence}. Annotators receive the same rendered input as the filtering judges and the evaluators under test (Figure~\ref{fig:ui}; guidelines in Appendix~\ref{app:prompt_annotator}), with trajectory slot positions randomized and target steering instructions $s_{c,t,\ell}$ hidden.

\subsection{Inter-Annotator Agreement}
\label{app:panel_alpha}

Table~\ref{tab:alpha} and Table~\ref{tab:panel} quantify inter-annotator reliability across the human annotation panel and compare human consistency against the synthetic ground truth label $\star$.

As shown in Table~\ref{tab:alpha}, evaluating Krippendorff's $\alpha$ across the human-only panel $(a_1, a_2, a_3)$ yields a baseline reliability of $+0.350$ on unfiltered data ($\mathcal{D}_0$), which increases slightly to $+0.383$ on fully filtered pairs ($\mathcal{D}_2$). Replacing a single human rater with the synthetic label $\star$ and averaging across panel permutations $(a_i, a_j, \star)$ substantially improves inter-rater reliability, raising Krippendorff's $\alpha$ to $+0.496$ on $\mathcal{D}_2$ ($\Delta = +0.113$). This increase indicates that the synthetic ground truth label provides a more consistent central consensus than individual human annotators.

Table~\ref{tab:panel} evaluates the reliability of the majority vote baseline $\mathrm{maj}$ across the 150 annotated pairs in $\mathcal{D}_0$. The overall panel achieves a pooled Krippendorff's $\alpha$ of $+0.350$. Exactly 51\% of pairs (77/150) achieve unanimous agreement (3--0 vote), while the remaining 49\% represent 2--1 split decisions. Pairwise inter-annotator agreement varies widely, with Cohen's $\kappa$ ranging from $+0.63$ down to $+0.12$ across panel configurations.

\begin{table}%
\caption{\textbf{Krippendorff's $\alpha$ over a panel of three raters.} The first row is the human panel $(a_1, a_2, a_3)$. The second row substitutes one human rater with synthetic label $\star$ and reports mean $\pm$ SD across permutations $(a_1, a_2, \star)$, $(a_1, a_3, \star)$, and $(a_2, a_3, \star)$. $\Delta$ represents the difference between hybrid and human-only panels.}
\label{tab:alpha}
\centering
\small
\begin{tabular}{@{}llll@{}}
\toprule
rater pool & $\mathcal{D}_0$ & $\mathcal{D}_1$ & $\mathcal{D}_2$ \\
\midrule
$(a_1, a_2, a_3)$                                & $+0.350$ & $+0.369$ & $+0.383$ \\
$(a_i, a_j, \star)$, mean $\pm$ SD & $+0.372 \pm 0.055$ & $+0.479 \pm 0.069$ & $\mathbf{+0.496} \pm 0.065$ \\
\midrule
$\Delta$                                         & $+0.022$ & $+0.110$ & $+0.113$ \\
\bottomrule
\end{tabular}
\end{table}

\begin{table}%
\caption{\textbf{Reliability metrics for majority vote reference $\mathrm{maj}$ across the 150 rated pairs in $\mathcal{D}_0$ (75 per panel).} Unanimous indicates a 3--0 vote. Lower blocks report pairwise agreement metrics for two-annotator panel subsets.}
\label{tab:panel}
\centering
\small
\begin{tabular}{@{}lrrr@{}}
\toprule
measure & pooled (150) & panel 1 (75) & panel 2 (75) \\
\midrule
Krippendorff's $\alpha$ & $\mathbf{+0.350}$ [+0.24, +0.45] & $+0.451$ [+0.29, +0.60] & $+0.240$ [+0.09, +0.38] \\
unanimous (3--0)        & \textbf{77/150 = 51\%} & 44/75 = 59\% & 33/75 = 44\% \\
\midrule
\multicolumn{4}{@{}l}{\emph{Agreement between every two annotators of a panel}} \\
& agreement & Cohen's $\kappa$ & both said ``certain'' \\
\midrule
panel 1, $(a_1, a_2)$ & 81.3\% (61/75) & $\mathbf{+0.63}$ & 85\% (28/33) \\
panel 1, $(a_1, a_3)$ & 65.3\% (49/75) & $+0.32$ & 75\% (24/32) \\
panel 1, $(a_2, a_3)$ & 70.7\% (53/75) & $+0.41$ & 83\% (30/36) \\
panel 2, $(a_1, a_2)$ & 66.7\% (50/75) & $+0.34$ & 81\% (13/16) \\
panel 2, $(a_1, a_3)$ & 65.3\% (49/75) & $+0.27$ & 68\% (15/22) \\
panel 2, $(a_2, a_3)$ & 56.0\% (42/75) & $\mathbf{+0.12}$ & 73\% (27/37) \\
\midrule
mean & \textbf{67.6\%} & $\mathbf{+0.35}$ & \\
\bottomrule
\end{tabular}
\end{table}

\subsection{Confidence and Task Difficulty}
\label{app:panel_confidence}

Table~\ref{tab:confidence} investigates whether filter judges select for human confidence or agreement. Annotators rate each pair on a three-point scale ($0$ for guess, $1$ for leaning, $2$ for certain), where the per-pair confidence rating is the mean of three annotator scores.

Human agreement correlates positively with confidence: Krippendorff's $\alpha$ reaches $+0.545$ on the 41 pairs where all three annotators indicate ``certain'', compared to $+0.270$ on the remaining 109 pairs where at least one annotator is unsure. However, filter judges do not select for higher annotator confidence. Mean confidence moves by only $+0.02$ on a $0$--$2$ scale between unfiltered $\mathcal{D}_0$ and dual-filtered $\mathcal{D}_2$. Furthermore, neither $J_1$ nor $J_2$ retains a significantly more confident pair subset than the one it removes ($p = 1.000$ for $J_1$, $p = 0.172$ for $J_2$). Given the sample sizes of removed pairs ($n = 36$ for $J_1$ and $n = 10$ for $J_2$), the minimum detectable confidence difference is approximately $0.20$.

These results indicate that filter judges discard pairs where annotators split on majority consensus rather than pairs annotators find inherently difficult. Consequently, filtering raises label agreement without rendering the underlying evaluation task trivial: mean annotator confidence remains $1.51$ out of $2.00$ on $\mathcal{D}_2$, reflecting non-trivial judgment calls for human evaluators.

\begin{table}%
\caption{\textbf{Human annotator confidence and agreement metrics across filter depths $\mathcal{D}_k$.} Per-pair confidence reflects the mean of three annotator ratings ($0 = \text{guess}$, $1 = \text{leaning}$, $2 = \text{certain}$). Statistical significance ($p$) is computed via permutation tests over 100{,}000 shuffles.
$\alpha$ is Krippendorff's $\alpha$.}
\label{tab:confidence}
\centering
\small
\begin{tabular}{@{}lrrr@{}}
\toprule
subset & $|\mathcal{D}|$ & mean confidence & SD \\
\midrule
$\mathcal{D}_0$ (unfiltered) & 150 & 1.49 & 0.43 \\
$\mathcal{D}_1$ ($J_1$)      & 114 & 1.49 & 0.44 \\
$\mathcal{D}_2$ ($J_1\&J_2$) & 104 & 1.51 & 0.44 \\
\midrule
\multicolumn{4}{@{}l}{\emph{Kept against removed, per filter judge}} \\
judge & kept & removed & difference ($p$) \\
\midrule
$J_1$ & 1.49 ($\mathcal{D}_1$, 114) & 1.49 ($\mathcal{D}_0 \setminus \mathcal{D}_1$, 36) & $\mathbf{+0.00}$ (1.000) \\
$J_2$ & 1.51 ($\mathcal{D}_2$, 104) & 1.30 ($\mathcal{D}_1 \setminus \mathcal{D}_2$, 10) & $+0.21$ (0.172) \\
\midrule
\multicolumn{4}{@{}l}{\emph{Agreement does track confidence, which the judges do not select for}} \\
subset of $\mathcal{D}_0$ & $|\mathcal{D}|$ & $\alpha$ & \\
\midrule
all three said ``certain'' & 41  & $\mathbf{+0.545}$ & \\
anyone was unsure          & 109 & $+0.270$ & \\
\bottomrule
\end{tabular}
\end{table}

\subsection{Sampling Robustness}
\label{app:panel_poststrat}

Table~\ref{tab:poststrat} verifies that the 150 human-annotated pairs accurately represent the full 1{,}000-pair dataset. In the human evaluation study, pairs are sampled uniformly across domain and criterion combinations. However, the full dataset features non-uniform domain allocations (e.g., airline comprises 13.5\% of the 1{,}000 pairs but 23.1\% of the annotated sample).

To evaluate potential sampling bias, we post-stratify synthetic label accuracy using actual population weights across four axes: domain, criterion, agent model, and level contrast. Re-weighting shifts reported synthetic label accuracy by less than 1.0\text{pt} across all configurations, well within the sampling standard error of $\pm 3.7\text{pt}$.

Furthermore, dataset retention rates under filtering in the human sample match full population proportions closely: the 150-pair sample survives $J_1$ at 76.0\% (114/150) and dual filtering at 69.3\% (104/150), compared to 76.4\% (859/1{,}125) and 68.4\% (770/1{,}125) across the full dataset. This confirms that the annotated subset is representative of overall pipeline behavior.

\begin{table}[H]
\caption{\textbf{Synthetic label accuracy (agreement with human majority vote) across raw sample estimates and population re-weighted strata ($\mathcal{D}_k$).} Post-stratification shifts headline accuracy by less than 1.0\text{pt}, confirming sample representativeness within a $\pm 3.7\text{pt}$ standard error.}
\label{tab:poststrat}
\centering
\small
\begin{tabular}{@{}lrrrrr@{}}
\toprule
subset & as reported & domain-weighted & criterion-weighted & agent-weighted & contrast-weighted \\
\midrule
$\mathcal{D}_0$ (unfiltered) & 73.3\% & 73.1\% & 73.3\% & 73.4\% & 73.7\% \\
$\mathcal{D}_1$ ($J_1$)      & 84.2\% & 83.4\% & 83.8\% & 84.6\% & 84.6\% \\
$\mathcal{D}_2$ ($J_1\&J_2$) & \textbf{84.6\%} & \textbf{83.9\%} & \textbf{84.2\%} & \textbf{85.0\%} & \textbf{85.3\%} \\
\bottomrule
\end{tabular}
\end{table}

\section{Filtering Gains and Human Alignment Dynamics}
\label{app:strata}

Table~\ref{tab:strata} breaks down filtering gains for agreement between synthetic labels $\star$ and human majority vote across four dataset axes. Filtering yields positive accuracy gains across all subsets, with the largest increases occurring where evaluation criteria are most difficult to judge without oversight. Specifically, communication clarity achieves a $+19$\text{pt} gain, though only 52\% of clarity pairs survive dual-filter verification. This indicates that the filtering pipeline discards a larger proportion of borderline trajectories to enforce the target criteria alignment. Task resolution yields a $+10$\text{pt} gain with a 72\% retention rate. Friendliness exhibits a modest $+3$\text{pt} gain with 84\% retention, starting from a high unfiltered baseline of 90\% and leaving minimal room for further optimization.

Two structural axes exhibit uniform gains across subsets. Across agent models, filtering improvements remain stable (+10\text{pt} to +13\text{pt}), indicating that the effect stems from dataset construction rather than specific architectures. Similarly, gains across level contrasts range tightly between +10\text{pt} and +12\text{pt}, proving filtering refines fine contrast pairs as effectively as wide ones without relying on gross trajectory contrasts.

The telecom domain is the sole exception, yielding a +1\text{pt} gain compared to +12\text{pt} to +17\text{pt} elsewhere. Table~\ref{tab:domains} suggests a structural mechanism: telecom features a 3{,}715-word policy, 30 user-side tools, a median length of 48 messages, and tool errors in 48\% of simulations, leaving filter judges with the least clear signal. Alternatively, telecom starts with the highest unfiltered accuracy at 79\%, leaving minimal room to gain. Cells hold 24 to 50 pairs throughout, supporting reliable subset ordering.

Table~\ref{tab:judge2} summarizes the second filter ($J_2$) performance. $J_2$ removes 10 of the 114 pairs passed by $J_1$, yielding a marginal $+0.4$\text{pt} gain in label purity at the cost of a $-7$\text{pt} drop in total yield. Human annotators support $J_1$ filter overrulings in 61\% of cases (22/36), but only 20\% (2/10) for $J_2$. Among the 10 pairs removed by $J_2$, the ground truth label $\star$ is human-verified as correct in 8 instances.
At this sample size, the second filter stage yields diminishing returns.
It incurs notable yield loss without providing a statistically meaningful improvement in label purity or alignment quality.
The full dataset retains depth $\mathcal{D}_2$, with ablation metrics reported for completeness.

\begin{table}%
\caption{\textbf{Agreement between synthetic labels $\star$ and human majority vote across filter depths $\mathcal{D}_k$ and dataset subsets.}
Gain reflects the accuracy change from two filters ($\mathcal{D}_2$) versus zero filters ($\mathcal{D}_0$); retention represents the proportion of target pairs retained after two filtering stages.}
\label{tab:strata}
\centering
\small
\begin{tabular}{@{}llrrrrr@{}}
\toprule
axis & subset & unfiltered ($\mathcal{D}_0$) & 1 filter ($\mathcal{D}_1$) & 2 filters ($\mathcal{D}_2$) & gain & retention \\
\midrule
\multicolumn{2}{@{}l}{\emph{By criterion}} \\
& friendliness           & 90\% (45/50) & 93\% (42/45) & 93\% (39/42) & $+3$pt  & 84\% \\
& task resolution        & 68\% (34/50) & 77\% (30/39) & 78\% (28/36) & $+10$pt & 72\% \\
& communication clarity  & 62\% (31/50) & 80\% (24/30) & 81\% (21/26) & $+19$pt & 52\% \\
\midrule
\multicolumn{2}{@{}l}{\emph{By domain}} \\
& airline & 75\% (27/36) & 92\% (23/25) & 92\% (22/24) & $+17$pt & 67\% \\
& retail  & 67\% (26/39) & 80\% (24/30) & 84\% (21/25) & $+17$pt & 64\% \\
& banking & 72\% (26/36) & 85\% (22/26) & 84\% (21/25) & $+12$pt & 69\% \\
& telecom & 79\% (31/39) & 82\% (27/33) & 80\% (24/30) & $+1$pt & 77\% \\
\midrule
\multicolumn{2}{@{}l}{\emph{By agent model}} \\
& gpt-oss-20b        & 76\% (35/46) & 90\% (27/30) & 89\% (24/27) & $+13$pt & 59\% \\
& nemotron-lightning-3.5 & 71\% (36/51) & 81\% (34/42) & 83\% (29/35) & $+12$pt & 69\% \\
& gpt-oss-120b       & 74\% (39/53) & 83\% (35/42) & 83\% (35/42) & $+10$pt & 79\% \\
\midrule
\multicolumn{2}{@{}l}{\emph{By level contrast}} \\
& bad--good (widest) & 80\% (37/46) & 90\% (35/39) & 92\% (33/36) & $+11$pt & 78\% \\
& ok--good           & 75\% (38/51) & 86\% (31/36) & 87\% (26/30) & $+12$pt & 59\% \\
& bad--ok            & 66\% (35/53) & 77\% (30/39) & 76\% (29/38) & $+10$pt & 72\% \\
\bottomrule
\end{tabular}
\end{table}

\begin{table}%
\caption{\textbf{Domain complexity metrics across evaluation environments.} Tool name overlap between domain pairs ranges from 0.03 to 0.11 by Jaccard index; banking represents the only domain featuring active retrieval.}
\label{tab:domains}
\centering
\small
\begin{tabular}{@{}lrrrrrrrr@{}}
\toprule
domain & \shortstack[r]{policy\\words} & \shortstack[r]{agent\\tools} & \shortstack[r]{user\\tools} & sims & \shortstack[r]{median\\msgs} & \shortstack[r]{median\\tool calls} & \shortstack[r]{tool-error\\sims} & \shortstack[r]{mean\\reward} \\
\midrule
airline & 1{,}313 & 14 & 0 & 440 & 18 & 5 & 21\% & 0.434 \\
banking & 926 & 16 & 2 & 855 & 30 & 7 & 3\% & \textbf{0.037} \\
retail  & 1{,}158 & 16 & 0 & 980 & 22 & 6 & 34\% & 0.353 \\
telecom & \textbf{3{,}715} & 13 & \textbf{30} & 1{,}072 & \textbf{48} & \textbf{10} & \textbf{48\%} & 0.243 \\
\bottomrule
\end{tabular}
\end{table}

\begin{table}%
\caption{\textbf{Trade-off analysis for the second-stage filter $J_2$.} Among the 10 pairs eliminated by stage two, human majority vote supports the original ground truth label in 8 cases.}
\label{tab:judge2}
\centering
\small
\begin{tabular}{@{}lrrlr@{}}
\toprule
human panel & pairs removed & label was right & purity & yield \\
\midrule
1 & 6  & 4 & 86.0 $\rightarrow$ 88.2\% ($+2.3$pt) & $-8$pt \\
2 & 4  & 4 & 82.5 $\rightarrow$ 81.1\% ($-1.3$pt) & $-5$pt \\
\midrule
pooled & \textbf{10} & \textbf{8} & \textbf{84.2 $\rightarrow$ 84.6\% ($\mathbf{+0.4}$pt)} & $\mathbf{-7}$\textbf{pt} \\
\bottomrule
\end{tabular}
\end{table}

\section{The Full Evaluator Sweep and Contamination Diagnostic}
\label{app:evaluators}

\subsection{The Full Sweep}
\label{app:evaluators_all}

Table~\ref{tab:evaluators_full} reports performance metrics across all 25 evaluators, serving as the unabridged version of Table~\ref{tab:evaluators} in the main text (which presents a 12-evaluator subset selected to span accuracy ranges and model families). All summary statistics, subset means, and correlation analyses in Section~\ref{sec:discriminates} (including Table~\ref{tab:correlations}) are computed over the full 25-evaluator roster. Evaluator accuracy is measured as $\mathrm{Agr}(f, \star; \mathcal{D})$ across all 1{,}000 pairs in $\mathcal{D}_2$ with ties treated as incorrect, reported as the mean $\pm$ SD across three independent runs. Accuracy spans a 48.3\text{pt} range, from \texttt{claude-opus-5} at 85.1\% down to \texttt{qwen3-4b} at 36.8\%.

Table~\ref{tab:evaluators_full} introduces two additional column blocks omitted from the main text due to space constraints: accuracy breakdown by level contrast and accuracy breakdown by agent model. Similar to the domain and criterion blocks, each of these blocks partitions the 1{,}000 pairs such that their weighted average equals the total dataset accuracy. Row indicators denote specific evaluator conditions: $\dagger$ identifies the two models where parse failures are marked as incorrect, while $*$ marks the three models utilized in dataset construction (analyzed further in Appendix~\ref{app:contamination}).

\begin{sidewaystable}
\caption{\textbf{The complete sweep: all 25 evaluators across 1{,}000 data pairs, repeat $n = 3$.}
Every statistic reported in Section~\ref{sec:discriminates} is computed over these 25 rows.
Table~\ref{tab:evaluators} in the main paper shares the row numbering.
Leniency is the mean of every score the evaluator emits.
It strongly correlates with accuracy (Table~\ref{tab:correlations}).
Level contrast indicates the two steering levels each trajectory pair has: \textsf{b--o} is \textsf{bad}--\textsf{ok} ($n = 313$), \textsf{o--g} is \textsf{ok}--\textsf{good} ($n = 328$), \textsf{b--g} is \textsf{bad}--\textsf{good} ($n = 359$).
Agent model columns are the three trajectory generators of Table~\ref{tab:axes}: \textsf{oss-20} is \texttt{gpt-oss-20b} ($n = 325$), \textsf{oss-120} is \texttt{gpt-oss-120b} ($n = 334$), \textsf{nemo.} is \texttt{nemotron-lightning-3.5} ($n = 341$).
Note that agent model is a property of the data point being judged rather than of the evaluator.
The final row is the unweighted mean over all 25 evaluators.
Released is the public release month.
Temperature = 0 except where unavailable$^{\ddagger}$.
Preference is derived from individual trajectory score gaps; ties count as incorrect.
Accuracy and score denote benchmark-level accuracy and raw score output of the evaluators, respectively.
SD is standard deviation.
$^{*}$Contaminated by self-assessment bias because the model is used during data synthesis (Appendix~\ref{app:contamination}).
$^{\dagger}$Parse failures on 192 and 119 calls after repeated retries, counted incorrect.
\textbf{Bolded} numbers indicate best performance within the column.}
\label{tab:evaluators_full}
\centering
\footnotesize
\setlength{\tabcolsep}{2pt}
\begin{tabular}{@{}rlrrrccrrrrrrrrrrrrrrrrr@{}}
\toprule
& & & \multicolumn{2}{c}{parameters} & & accuracy (\%) & & average & tie & \multicolumn{4}{c}{by domain (\%)} & \multicolumn{3}{c}{by criterion (\%)} & \multicolumn{3}{c}{by level contrast (\%)} & \multicolumn{3}{c}{by agent model (\%)} & distinct \\
\cmidrule(lr){4-5}\cmidrule(lr){11-14}\cmidrule(lr){15-17}\cmidrule(lr){18-20}\cmidrule(lr){21-23}
\# & evaluator & released & total & active & reas. & mean $\pm$ SD & leniency & score SD & (\%) & airl. & bank. & retail & telec. & clar. & friend. & task & b--o & o--g & b--g & oss-20 & oss-120 & nemo. & scores \\
\midrule
1 & \texttt{claude-opus-5} & 2026-07 & closed & closed & Yes & \textbf{85.1} $\pm$ 0.93 & 0.324 & 0.023 & 4.1 & \textbf{84} & \textbf{82} & \textbf{86} & \textbf{87} & 73 & 96 & \textbf{85} & 85 & \textbf{82} & \textbf{88} & \textbf{85} & 88 & \textbf{82} & 69 \\
2 & \texttt{claude-sonnet-5} & 2026-06 & closed & closed & Yes & 83.8 $\pm$ 0.72 & 0.458 & 0.040 & 5.8 & 82 & 81 & 84 & \textbf{87} & 73 & 95 & 82 & 85 & 81 & 86 & 81 & \textbf{89} & \textbf{82} & 34 \\
3 & \texttt{kimi-k3} & 2026-07 & 2.8T & 104B & Yes & 83.6 $\pm$ 0.12 & 0.508 & 0.036 & 6.0 & 83 & 80 & 84 & \textbf{87} & 71 & \textbf{97} & 81 & \textbf{87} & 78 & 86 & 82 & 87 & 81 & 37 \\
4 & \texttt{gpt-5.6-terra}$^{\ddagger}$ & 2026-07 & closed & closed & Yes & 81.7 $\pm$ 0.49 & 0.560 & 0.043 & \textbf{3.4} & 81 & 81 & 82 & 82 & 69 & 96 & 78 & 86 & 75 & 84 & 79 & 86 & 80 & 88 \\
5 & \texttt{gpt-5.6-luna}$^{\ddagger}$ & 2026-07 & closed & closed & Yes & 81.6 $\pm$ 0.64 & 0.563 & 0.045 & 3.8 & 82 & 81 & 81 & 83 & 69 & 93 & 81 & \textbf{87} & 75 & 83 & 81 & 85 & 79 & 83 \\
6 & \texttt{glm-5p2} & 2026-06 & 753B & $\sim$30B & Yes & 81.5 $\pm$ 0.50 & 0.521 & 0.051 & 10.1 & \textbf{84} & 81 & 81 & 82 & \textbf{78} & 92 & 73 & 81 & 78 & 85 & 80 & 85 & 79 & 31 \\
7 & \texttt{gpt-oss-120b}$^{*}$ & 2025-08 & 116.8B & 5.1B & Yes & 80.5 $\pm$ 0.40 & 0.624 & 0.050 & 9.9 & 80 & 81 & 80 & 80 & \textbf{78} & 87 & 75 & 83 & 74 & 84 & 78 & 87 & 77 & 45 \\
8 & \texttt{gpt-5.4-nano}$^{\ddagger}$ & 2026-03 & closed & closed & No & 79.4 $\pm$ 1.16 & 0.631 & 0.052 & 7.8 & 81 & 78 & 78 & 81 & 74 & 84 & 80 & 82 & 72 & 83 & 80 & 84 & 75 & 61 \\
9 & \texttt{gpt-5.4-mini}$^{\ddagger}$ & 2026-03 & closed & closed & No & 79.4 $\pm$ 0.79 & 0.628 & 0.053 & 6.1 & 74 & 79 & 80 & 81 & 71 & 91 & 75 & 85 & 72 & 82 & 79 & 84 & 76 & 89 \\
10 & \texttt{gpt-5.6-sol}$^{\ddagger}$ & 2026-07 & closed & closed & Yes & 78.7 $\pm$ 0.40 & 0.543 & 0.035 & 4.9 & 79 & 77 & 81 & 77 & 64 & \textbf{97} & 74 & 84 & 72 & 80 & 77 & 82 & 77 & 86 \\
11 & \texttt{claude-haiku-4-5} & 2025-10 & closed & closed & No & 78.7 $\pm$ \textbf{0.10} & 0.505 & \textbf{0.000} & 11.6 & 81 & 79 & 76 & 81 & 76 & 85 & 74 & 75 & 77 & 84 & 76 & 85 & 75 & 32 \\
12 & \texttt{qwen3p8-max} & 2026-08 & 2.4T & 95B & Yes & 78.0 $\pm$ 0.36 & 0.521 & 0.037 & 11.7 & 74 & 78 & 77 & 81 & 70 & 94 & 68 & 81 & 73 & 80 & 76 & 82 & 76 & 30 \\
13 & \texttt{nemotron-3-ultra} & 2026-06 & 549B & 55B & Yes & 76.8 $\pm$ 0.76 & 0.558 & 0.044 & 14.2 & 75 & 80 & 70 & 81 & 71 & 88 & 71 & 79 & 70 & 81 & 76 & 81 & 73 & 28 \\
14 & \texttt{llama3.1-70b} & 2024-07 & 70.6B & dense & No & 70.0 $\pm$ 0.55 & 0.632 & 0.016 & 22.8 & 67 & 73 & 67 & 72 & 64 & 69 & 77 & 73 & 61 & 76 & 73 & 79 & 58 & 14 \\
15 & \texttt{gemini-3.7-flash} & 2026-08 & closed & closed & Yes & 69.5 $\pm$ 0.45 & 0.615 & 0.026 & 23.4 & 71 & 70 & 69 & 69 & 57 & 95 & 54 & 72 & 64 & 73 & 69 & 74 & 66 & 34 \\
16 & \texttt{gpt-oss-20b}$^{*}$ & 2025-08 & 20.9B & 3.6B & Yes & 68.7 $\pm$ 1.01 & 0.625 & 0.056 & 21.7 & 71 & 70 & 68 & 67 & 74 & 80 & 51 & 69 & 67 & 70 & 68 & 74 & 64 & 31 \\
17 & \texttt{gemma-4-31b} & 2026-03 & 32.2B & dense & Yes & 68.1 $\pm$ 0.15 & 0.657 & 0.023 & 27.4 & 63 & 67 & 66 & 74 & 47 & 92 & 63 & 75 & 56 & 73 & 71 & 75 & 59 & 14 \\
18 & \texttt{deepseek-v4-pro} & 2026-08 & 1.6T & 49B & Yes & 67.0 $\pm$ 0.67 & 0.505 & 0.045 & 20.4 & 62 & 70 & 69 & 65 & 61 & 90 & 49 & 69 & 63 & 69 & 63 & 70 & 68 & 23 \\
19 & \texttt{mistral-large2} & 2024-07 & 123B & dense & No & 64.8 $\pm$ 0.40 & 0.717 & 0.014 & 28.1 & 59 & 63 & 62 & 72 & 57 & 63 & 74 & 71 & 53 & 70 & 70 & 72 & 53 & 12 \\
20 & \texttt{nemotron-lightning-3.5}$^{*}$ & 2026-08 & 32B & 3B & Yes & 62.4 $\pm$ 0.93 & 0.705 & 0.060 & 28.6 & 61 & 58 & 60 & 69 & 47 & 75 & 64 & 67 & 52 & 68 & 66 & 69 & 52 & 22 \\
21 & \texttt{gemini-3.5-flash-lite} & 2026-07 & closed & closed & No & 56.1 $\pm$ 0.68 & 0.723 & 0.055 & 36.3 & 53 & 58 & 47 & 65 & 52 & 59 & 58 & 63 & 41 & 64 & 64 & 62 & 44 & 16 \\
22 & \texttt{llama4-maverick} & 2025-04 & 400B & 17B & No & 55.6 $\pm$ 1.56 & 0.765 & 0.039 & 35.9 & 47 & 56 & 54 & 61 & 52 & 55 & 60 & 58 & 51 & 58 & 61 & 60 & 45 & 15 \\
23 & \texttt{llama3.1-8b}$^{\dagger}$ & 2024-07 & 8.0B & dense & No & 43.1 $\pm$ 0.47 & 0.622 & 0.005 & 43.7 & 43 & 46 & 48 & 35 & 28 & 43 & 57 & 44 & 39 & 47 & 42 & 46 & 42 & 12 \\
24 & \texttt{qwen3-1p7b}$^{\dagger}$ & 2025-04 & 2.0B & dense & Yes & 42.8 $\pm$ 0.35 & 0.760 & 0.076 & 31.9 & 44 & 44 & 42 & 42 & 43 & 47 & 38 & 47 & 33 & 48 & 43 & 45 & 40 & 15 \\
25 & \texttt{qwen3-4b} & 2025-08 & 4.4B & dense & No & 36.8 $\pm$ 0.65 & 0.804 & 0.029 & 56.9 & 36 & 34 & 32 & 44 & 30 & 25 & 55 & 43 & 27 & 40 & 43 & 41 & 26 & 17 \\
\midrule
\multicolumn{6}{@{}l}{\emph{Mean, all 25 evaluators}} & 70.1 &  &  &  & 68.7 & 69.9 & 69.0 & 72.2 & 62.0 & 79.5 & 67.9 & 73.2 & 63.4 & 73.6 & 70.5 & 74.9 & 65.1 &  \\
\bottomrule
\end{tabular}
\end{sidewaystable}

\subsection{Contamination Diagnostic}
\label{app:contamination}

Three evaluators in our sweep contribute directly to benchmark construction: \texttt{nemotron-lightning-3.5} serves as filter judge $J_1$, \texttt{gpt-oss-120b} serves as $J_2$, and these two models in addition to \texttt{gpt-oss-20b} generate 325--341 of the evaluated trajectory pairs. While filter judge influence is intrinsic to dataset definition, generation influence can be isolated. Table~\ref{tab:contamination} evaluates potential self-preference bias by comparing evaluator accuracy on self-generated trajectories versus external trajectories.

The diagnostic indicates that self-preference is not systematic. Only \texttt{gpt-oss-120b} exhibits higher accuracy on self-generated outputs ($+9.1\text{pt}$), \texttt{gpt-oss-20b} displays negligible shift ($-0.6\text{pt}$), and \texttt{nemotron-lightning-3.5} performs $-15.7\text{pt}$ on its own trajectories. This result is primarily driven by trajectory difficulty shifts: because agent assignment is deterministic per cell rather than random, excluding self-generated pairs confounds underlying trajectory difficulty with evaluator bias. As reported in Table~\ref{tab:evaluators_full} under the ``by agent model'' block, these trajectory difficulty shifts are revealed in the mean accuracy across all 25 evaluators: $74.9\%$ on \texttt{gpt-oss-120b} trajectories, $70.5\%$ on \texttt{gpt-oss-20b} trajectories, and drops to $65.1\%$ on \texttt{nemotron-lightning-3.5} trajectories. The observed gaps in Table~\ref{tab:contamination} closely track benchmark-wide trajectory difficulty shifts rather than systematic self-preference bias.

Notably, all three generator models belong to the sub-7B active parameter MoE tier, highlighting a parameter tier constraint when evaluating lightweight model architectures.

\begin{table}%
\caption{\textbf{Contamination diagnostic comparing evaluator accuracy on self-generated versus external trajectory pairs.} Values represent means across three independent evaluation runs.}
\label{tab:contamination}
\centering
\small
\begin{tabular}{@{}lrrr@{}}
\toprule
& \multicolumn{2}{c}{accuracy (\%)} & \\
\cmidrule(lr){2-3}
evaluator & its own pairs & the other pairs & gap (pt) \\
\midrule
\texttt{gpt-oss-120b}       & 86.5 ($n = 334$) & 77.4 ($n = 666$) & $\mathbf{+9.1}$ \\
\texttt{gpt-oss-20b}        & 68.3 ($n = 325$) & 68.9 ($n = 675$) & $-0.6$ \\
\texttt{nemotron-lightning-3.5} & 52.1 ($n = 341$) & 67.8 ($n = 659$) & $\mathbf{-15.7}$ \\
\bottomrule
\end{tabular}
\end{table}

\section{Cost and Caching}
\label{app:cost}

The dataset curation pipeline generates 1{,}000 validated trajectory pairs at a total cost of \$23.63, or \$0.024 per kept pair (Table~\ref{tab:cost_overhead}). Steering instruction generation and filter judges account for 5.0\% of total token volume and 12.4\% of dollar costs.

\begin{table}%
\caption{\textbf{Computational overhead and curation cost.} All costs are computed or estimated with the model service provider's token-based pricing model.}
\label{tab:cost_overhead}
\centering
\small
\begin{tabular}{@{}lrrrrr@{}}
\toprule
\textbf{Pipeline Stage} & \textbf{API Calls} & \textbf{Tokens} & \textbf{Token Share} & \textbf{Uncached Cost} & \textbf{Effective Cost} \\
\midrule
1. Steering Generation   & 1,125  & 2.56M  & 0.3\%  & \$0.85  & \$0.67  \\
2. Trajectory Generation  & 77,980 & 709.4M & 95.0\% & \$51.16 & \$18.36 \\
3. Filter Judges ($K=2$) & 2,879  & 34.8M  & 4.7\%  & \$4.42  & \$2.27  \\
\midrule
\textbf{Total Pipeline}  & \textbf{81,984} & \textbf{746.7M} & \textbf{100.0\%} & \textbf{\$56.43} & \textbf{\$23.63} \\
\bottomrule
\end{tabular}
\end{table}

Prompt caching is the primary driver of generation cost reduction.
In an agentic loop, the full conversation history is re-sent at each turn, making almost every request carry a prefix the server already holds.
Across the generation run, 98\% of all processed tokens are prompt tokens.

Table~\ref{tab:cache} reports the prompt cache hit rate across all 77{,}980 trajectory calls. The overall hit rate reaches 90.5\%, ranging from 87.9\% to 95.4\% across individual models. This caching efficiency reduces total API costs by a factor of two to four depending on model pricing.

\begin{table}%
\caption{\textbf{Prompt cache hit rates across all 77{,}980 trajectory generation calls.}}
\label{tab:cache}
\centering
\small
\begin{tabular}{@{}lrrr@{}}
\toprule
model & prompt tokens & cached & rate \\
\midrule
\texttt{nemotron-lightning-3.5}     & 350{,}084{,}755 & 312{,}862{,}256 & 89.4\% \\
\texttt{gpt-oss-20b}                & 137{,}504{,}779 & 125{,}776{,}651 & 91.5\% \\
\texttt{qwen3-30b-a3b-instruct}     & 107{,}519{,}458 & 102{,}523{,}608 & 95.4\% \\
\texttt{gpt-oss-120b}               & 98{,}571{,}871  & 86{,}652{,}136  & 87.9\% \\
\midrule
total & \textbf{693{,}680{,}863} & \textbf{627{,}814{,}651} & \textbf{90.5\%} \\
\bottomrule
\end{tabular}
\end{table}

\section{Prompts}
\label{app:prompts}

This section reproduces every pipeline prompt verbatim in execution order: shared criterion definitions (Appendix~\ref{app:prompt_criteria}), steering instruction generation (Appendix~\ref{app:prompt_generator}), agent system prompt wrapping (Appendix~\ref{app:prompt_wrapper}), filter judging (Appendix~\ref{app:prompt_judge}), model evaluation (Appendix~\ref{app:prompt_evaluator}), and human annotation instructions (Appendix~\ref{app:prompt_annotator}). Variable names enclosed in braces (e.g., \texttt{\{criterion\_name\}}) represent runtime substitution slots. Agent and user simulator system prompts are retained directly from $\tau^3$-bench without modification; only the steering wrapper (Appendix~\ref{app:prompt_wrapper}) is appended to agent instructions.

\subsection{Criterion Definitions}
\label{app:prompt_criteria}

Below is the verbatim text for the three evaluation criteria $c$. The exact same criterion names and descriptions are supplied to the generator, filter judges, model evaluators, and human annotators, ensuring information parity across all pipeline components. Table~\ref{tab:axes} provides abridged summaries for reference.

\begin{lstlisting}
name: friendliness
description: Warmth and consideration toward the customer: whether the agent acknowledges their situation and how they feel about it, delivers unwelcome news with care, and leaves them feeling attended to. Judge the manner, not whether the request was resolved.

name: communication_clarity
description: How easily the customer can follow the agent: whether the main point is findable, whether technical or policy language is explained, whether multi-part information is organised, and whether the customer is left knowing what is true and what happens next. Judge the presentation, not the warmth or the outcome.

name: task_resolution
description: Whether the customer's actual problem was settled: did the agent establish what was needed, take the actions that would resolve it, and leave the customer with the outcome they came for. A correct refusal counts as resolution -- if the request was not permitted, saying so plainly and explaining why resolves it, while quietly doing it anyway does not. Judge the outcome, not the manner or how well it was explained.
\end{lstlisting}

\subsection{Steering Instruction Generator}
\label{app:prompt_generator}

The generator synthesizes three steering instructions $s_{c,t,\ell}$ for a given task cell $(t,c)$ in a single call, corresponding to levels $\ell \in \{\textsf{bad}, \textsf{ok}, \textsf{good}\}$. Task specifics (\texttt{\{task\_description\}} and \texttt{\{tool\_names\}}) are populated dynamically. Rule 5 prevents the generator from encoding unstated task outcomes, while Rule 1 ensures level contrasts reflect behavioral differences rather than varying degrees of task detail.

\begin{lstlisting}
You are designing an experiment about how well automatic evaluators judge the
quality of AI agent behaviour.

For ONE specific task and ONE specific quality metric, write three system-prompt
instructions that steer an agent to perform at three levels on that metric: BAD, OK
and GOOD.

## The metric

Name: {criterion_name}
What it means: {criterion_description}

## The task

What the user is trying to get done:
{task_description}

Tools the agent can use: {tool_names}

## Rules for the instructions you write

1. BE SPECIFIC TO THE KIND OF WORK THIS TASK INVOLVES -- NOT TO ITS PARTICULARS.
   Generic advice that would fit any task is one failure mode. Naming the task's
   details is the other, and it is worse. Do refer to which tools matter here, what
   kind of information the user needs, and the shape this interaction will take. Do
   NOT restate the specifics you were given: no names, no user or account or
   booking identifiers, no retelling of this person's circumstances. The agent
   learns those from the conversation itself. Putting them in its system prompt
   hands it information out of band, and makes your three instructions differ by
   how much detail they carry rather than by the metric.
   Wrong: "You are assisting Emma Kim (user ID emma_kim_9957) with cancelling
   reservation EHGLP3, and she was out of town relying on prior insurance."
   Right: "When the customer explains why they are asking, acknowledge the
   circumstances they raise before you get to the outcome."
2. MAKE THE THREE LEVELS SEPARATE. Someone reading the three resulting transcripts
   should be able to rank them on this metric without being told which is which. If
   two of your instructions would produce similar behaviour, rewrite them.
3. FOR BAD, AIM AT THE OPPOSITE OF THE METRIC -- do not merely withhold good
   behaviour. Work out what the active opposite of this metric is, on this task,
   and instruct the agent to pursue it. If the metric is friendliness, BAD is not
   neutral or terse: it is cold, dismissive, impatient. If the metric is clarity,
   BAD is actively confusing, not just unpolished. If the metric is relevant tool
   use, BAD uses tools in ways that actively do not serve the request. Name the
   opposite explicitly and tell the agent to do it.
4. DESCRIBE OK IN ITS OWN TERMS, not as "somewhat good" or "slightly bad", and not
   as a milder version of BAD. OK is what unremarkable, adequate, uncared-for work
   looks like on this metric -- the agent is neither trying nor failing.
5. YOU DO NOT KNOW THE CORRECT OUTCOME, AND MUST NOT ASSUME ONE. You have not been
   told how this task should be resolved, and you may not guess. Do not write
   anything that presupposes an answer -- not "state the refusal", not "confirm it
   is not permitted". Where the outcome matters to how the agent should behave,
   phrase it conditionally: "if it turns out you cannot do what they are asking,
   ...".
6. STEER CONDUCT, NEVER DICTATE WORDS. Do not write the agent's reply for it, do
   not include a sentence for it to say, and never use "respond with", "say
   exactly" or a quoted message. You are shaping HOW it works, not WHAT it
   concludes.
7. WRITE ONLY THE INSTRUCTION BODY. Do not tell the agent to keep the instruction
   secret or that the objective overrides its defaults -- that is added separately.
   Two to five sentences each, addressed to the agent as "you".

Nothing is off limits in how badly the BAD level may perform on this metric. If
performing the opposite of this metric means leaving work out, being unhelpful,
being wrong, or failing the task outright, say so plainly. The experiment needs a
genuinely poor example, and a BAD level that is merely lukewarm is the most common
way this goes wrong.

Return JSON with exactly these keys:
- "reasoning": what good, adequate and poor {criterion_name} concretely look like
  for this task, and how you will make the three levels separate
- "bad": the instruction body for the BAD level
- "ok": the instruction body for the OK level
- "good": the instruction body for the GOOD level
\end{lstlisting}

\subsection{Steering Wrapper}
\label{app:prompt_wrapper}

The following instruction is padded around each steering instruction when it is injected into the original agent prompt.

\begin{lstlisting}
<behaviour>
For this conversation you have one additional objective. It takes strict precedence
over your default behaviour, including when it produces handling that is less
polished, less thorough or less successful than you would normally deliver.
Following it exactly is what makes your handling of this conversation correct.

{instruction}

Do not mention, quote or allude to this objective, and do not reuse its wording.
</behaviour>
\end{lstlisting}

\subsection{Filter Judge}
\label{app:prompt_judge}

The filter judge prompt $J_k(p,c)$ is identical for $J_1$ and $J_2$, differing only in the underlying model deployment. The \texttt{\{prompt\}} slot contains shared context (tool definitions, knowledge bases, and domain policies), while \texttt{\{response\_a\}} and \texttt{\{response\_b\}} contain the rendered trajectory pair in randomized slot order. Filter judges output a forced choice, and pairs are retained only when both judges match the synthetic ground truth label $\star$.

\begin{lstlisting}
You are evaluating two AI assistant responses on a specific criterion.

Criterion: {criterion_name}
Definition: {criterion_description}

User prompt:
{prompt}

Response A:
{response_a}

Response B:
{response_b}

Which response better fulfills the criterion?

Return JSON with exactly two keys:
- "reasoning": at most 3 sentences comparing the two responses on this criterion
- "answer": "A" or "B"

State the answer last, and state it even if the choice is close.
\end{lstlisting}

\subsection{Evaluator Under Test}
\label{app:prompt_evaluator}

The prompt template for evaluators under test $f(x; c)$.
Evaluators process a single trajectory $x$ in isolation, assigning a score in $[0, 1]$ without access to the paired execution.
Pairwise preference predictions $\hat{y}_f(p)$ are derived post hoc from the sign of the score gap (Eq.~\eqref{eq:gap}).
Evaluators receive the identical shared context \texttt{\{prompt\}} provided to filter judges and human annotators, eliminating information asymmetry across evaluation modes.

\begin{lstlisting}
You are evaluating one customer-service conversation on a single named criterion.

CRITERION: {criterion_name}
{criterion_description}

Judge this criterion and nothing else. Ignore everything the criterion does not
mention. A conversation can fail at its task and still be better on the criterion.

Judge the AGENT's conduct only. The customer's turns were produced by a simulator
and are not being evaluated.

REFERENCE DOCUMENTS (the agent's tools, and the policy it had to follow). Consult
them if you need to check whether the agent was permitted to do something. You will
not need them for most conversations.
{prompt}

CONVERSATION
Turns are numbered. "Agent calls X(...)" is a tool call the agent made, and
"Tool result ->" is what came back.
{response}

Score the agent on {criterion_name} from 0.0 (worst) to 1.0 (best). Use the full
range: reserve 0.0 and 1.0 for clear cases and place ordinary ones in between.

Return JSON with exactly two keys:
- "reasoning": at most 3 sentences, citing what in the conversation drove the score
- "score": a number between 0.0 and 1.0
\end{lstlisting}

\subsection{Human Annotator Instructions}
\label{app:prompt_annotator}

The instructions of the human annotation interface (Figure~\ref{fig:ui}). Annotators review these instructions prior to their initial trial and retain permanent access during annotation. On-screen confidence levels ($0$, $50$, and $100$) are mapped to numerical values ($0$, $1$, and $2$) as used throughout Appendix~\ref{app:panel_confidence}.
\begin{lstlisting}
Before you start
Pairwise annotation of agent conversations. Read this once.

Judge one criterion, and nothing else
Each pair is judged on a single named criterion.
Only that criterion counts. Ignore everything else. In an extreme scenario, a
conversation can fail the task and still be the better one on the metric.

The flow -- three steps per pair
1. Pick a side. Click the conversation that better fits the criterion, or press
   the left / right arrow key.
2. Rate your confidence: 1 for 0 (a guess), 2 for 50 (leaning one way), 3 for
   100 (certain).
3. Press Next (Enter) to submit and move on.
Backspace goes back to revise.

There is more to read if you need it
Under the header are buttons labelled "Additional documents to help you decide":
the agent's tool list, the policy it had to follow, and the knowledge base where
one exists. They are closed by default and open on a click, one at a time. Tool
results and searches inside a conversation open the same way.
You do not need any of it for most pairs. Reach for the policy when the question
is whether the agent was allowed to do something.

How much to deliberate
Use your best judgment, but it might hurt to think too much. In my experience
giving each example about a minute should be sufficient.
If the two really are hard to separate, still pick a side -- then rate it 0. That
records the pair honestly as a coin-flip instead of hiding a guess among your
real judgments.

    [ Start annotating ]

---- shown once at the start of each criterion's block of pairs ----
You are now judging
<criterion name>
<criterion description>
Press any key or click to begin this section

---- the standing question in the header of every pair ----
Which conversation is better on this criterion?
\end{lstlisting}

\section{Assets, Licenses, and Terms of Use}
\label{app:licenses}

Table~\ref{tab:assets} lists every existing asset this work builds on, its role in the pipeline, and its license and terms of use.
All open-weight models are accessed through a single hosted inference provider rather than downloaded, and all proprietary models are accessed through their vendors' paid APIs, so each model is used under both its own license and the serving provider's terms of service.

\begin{table}[H]
\caption{\textbf{Existing assets used in this work.} Model roles refer to the pipeline stages of Table~\ref{tab:roles} and the evaluator sweep of Section~\ref{sec:discriminates}.}
\label{tab:assets}
\centering
\scriptsize
\setlength{\tabcolsep}{3pt}
\begin{tabular}{@{}>{\raggedright\arraybackslash}p{3.0cm}>{\raggedright\arraybackslash}p{3.7cm}>{\raggedright\arraybackslash}p{3.3cm}>{\raggedright\arraybackslash}p{3.3cm}@{}}
\toprule
asset & role in this work & license & terms of use \\
\midrule[\heavyrulewidth]
\multicolumn{4}{@{}l}{\emph{Benchmark substrate and serving infrastructure}} \\
\midrule[0.2pt]
$\tau^3$-bench, \texttt{sierra-research/} \texttt{tau2-bench} \citep{yao2025taubench, barres2026taubench, shi2026tauknowledge} & Task definitions, domain environments, tool sets, agent framework, and user simulator & MIT License & Free commercial and non-commercial use, modification, and redistribution provided the MIT notice (Copyright (c) 2025 Sierra Research) is preserved; evaluation runs are additionally subject to third-party LLM API terms \\
\midrule[0.2pt]
Fireworks AI serverless inference & Hosted inference for every open-weight model in the pipeline and the sweep & not applicable (service) & Fireworks AI terms of service \\
\midrule[\heavyrulewidth]
\multicolumn{4}{@{}l}{\emph{Open-weight models}} \\
\midrule[0.2pt]
\texttt{gpt-oss-20b}, \texttt{gpt-oss-120b} & Trajectory generation; \texttt{120b} also generates steering instructions and serves as filter judge $J_2$; both are evaluators & Apache 2.0 & model license and Fireworks AI terms of service \\
\midrule[0.2pt]
\texttt{nemotron-lightning-3.5} & Filter judge $J_1$, trajectory generation, and evaluator & NVIDIA OpenMDW-1.1 & as above \\
\midrule[0.2pt]
\texttt{nemotron-3-ultra} & Evaluator & NVIDIA OpenMDW-1.1 & as above \\
\midrule[0.2pt]
\texttt{qwen3-30b-a3b-instruct} & User simulator & Apache 2.0 & as above \\
\midrule[0.2pt]
\texttt{qwen3-1p7b}, \texttt{qwen3-4b} & Evaluators & Apache 2.0 & as above \\
\midrule[0.2pt]
\texttt{qwen3p8-max} & Evaluator & Apache 2.0 & as above \\
\midrule[0.2pt]
\texttt{llama3.1-8b}, \texttt{llama3.1-70b} & Evaluators & Llama 3.1 Community License & as above \\
\midrule[0.2pt]
\texttt{llama4-maverick} & Evaluator & Llama 4 Community License & as above \\
\midrule[0.2pt]
\texttt{gemma-4-31b} & Evaluator & Apache 2.0 & as above \\
\midrule[0.2pt]
\texttt{mistral-large2} & Evaluator & Mistral Research License & as above \\
\midrule[0.2pt]
\texttt{deepseek-v4-pro} & Evaluator & MIT License & as above \\
\midrule[0.2pt]
\texttt{glm-5p2} & Evaluator & MIT License & as above \\
\midrule[0.2pt]
\texttt{kimi-k3} & Evaluator & custom Kimi K3 license  & as above \\
\midrule[\heavyrulewidth]
\multicolumn{4}{@{}l}{\emph{Proprietary models (weights not released; API access only)}} \\
\midrule[0.2pt]
\texttt{claude-opus-5}, \texttt{claude-sonnet-5}, \texttt{claude-haiku-4-5} & Evaluators & closed weights, no public license & Anthropic API terms of service \\
\midrule[0.2pt]
\texttt{gpt-5.6-terra}, \texttt{gpt-5.6-luna}, \texttt{gpt-5.6-sol}, \texttt{gpt-5.4-nano}, \texttt{gpt-5.4-mini} & Evaluators & closed weights, no public license & OpenAI API terms of service \\
\midrule[0.2pt]
\texttt{gemini-3.7-flash}, \texttt{gemini-3.5-flash-lite} & Evaluators & closed weights, no public license & Google Gemini API terms of service \\
\bottomrule
\end{tabular}
\end{table}